\documentclass[journal]{IEEEtran}
\usepackage{graphics} 
\usepackage{graphicx}
\usepackage{multirow}
\usepackage{cite}
\usepackage{amsmath,amssymb,amsfonts}
\usepackage{algorithmic}
\usepackage{graphicx}
\usepackage{textcomp}
\usepackage{graphics} 

\usepackage{rotating}

\usepackage{epsfig} 
\usepackage{listings} 
\usepackage{multirow}
\usepackage{longtable}
\usepackage{graphicx}
\usepackage{textcomp}
\usepackage{rotating}
\usepackage{pdflscape}
\usepackage{afterpage}     
\usepackage{verbatim}
\usepackage{dcolumn}
\usepackage{adjustbox}
\usepackage[table,xcdraw]{xcolor}
\usepackage{url}
\usepackage[flushleft]{threeparttable}
\usepackage[caption=false]{subfig}
\usepackage{graphicx}
\usepackage{caption}
\usepackage{lipsum} 
\usepackage{array,multirow}
\usepackage[table,xcdraw]{xcolor}\usepackage{epsfig}

\def\BibTeX{{\rm B\kern-.05em{\sc i\kern-.025em b}\kern-.08em
    T\kern-.1667em\lower.7ex\hbox{E}\kern-.125emX}}
\begin{document}

\title{Micro Air Vehicle Link  (MAVLink) in a Nutshell: A Survey\thanks{The paper is accepted for publication in IEEE Access, June 2019.}}

\author{Anis~Koub\^aa,
Azza~Allouch,
Maram~Alajlan,
Yasir~Javed,
Abdelfettah~Belghith,
and~Mohamed~Khalgui, 

\thanks{A. Koub\^aa is with the Department of Computer Science, Prince Sultan
University, Riyadh PSU 12435, Saudi Arabia, also with CISTER/INESC TEC,
4200-135 Porto, Portugal, and also with ISEP-IPP, Porto, Portugal, (email: akoubaa@psu.edu.sa).}

\thanks{A. Allouch is with the School of Intelligent Systems Science and Engineering, Jinan University (Zhuhai Campus), Zhuhai 519070, China, also with the Faculty of Mathematical, Physical and Natural Sciences of Tunis, University of Tunis El Manar, Tunis 1068, Tunisia, and also
with the National Institute of Applied Sciences and Technology, University
of Carthage, Tunis 1080, Tunisia, (email: azza.allouch@coins-lab.org).}
\thanks{M. Alajlan is with the College of Computer and
Information Sciences, King Saud University, Riyadh, Saudi Arabia,
(email: maram.ajlan@coins-lab.org).}
\thanks{Y. Javed is with the College of Computer and Information
Sciences, Prince Sultan University, Riyadh, Saudi Arabia,
 (e-mail:
yjaved@psu.edu.sa).}

\thanks{A. Belghith is with the College of Computer and Information Sciences, King Saud University, Riyadh, Saudi Arabia, (e-mail:
abelghith@ksu.edu.sa).}
\thanks{M. Khalgui is with the School of Intelligent Systems Science
and Engineering, Jinan University (Zhuhai Campus), Zhuhai 519070,
China, and also with the National Institute of Applied Sciences
and Technology, University of Carthage, Tunis 1080, Tunisia, (e-mail:
khalgui.mohamed@gmail.com).}}

\markboth{Journal of \LaTeX\ Class Files,~Vol.~14, No.~8, August~2015}%
{Shell \MakeLowercase{\textit{et al.}}: Bare Demo of IEEEtran.cls for IEEE Journals}

\maketitle
\begin{abstract}

The Micro Air Vehicle Link (MAVLink in short) is a communication protocol for unmanned systems (e.g., drones, robots). It specifies a comprehensive set of messages exchanged between unmanned systems and ground stations. This protocol is used in major autopilot systems, mainly ArduPilot and PX4, and provides powerful features not only for monitoring and controlling unmanned systems missions, but also for their integration into the Internet. However, there is no technical survey and/or tutorial in the literature that presents these features or explains how to make use of them. Most of the references are online tutorials and basic technical reports, and none of them presents a comprehensive and a systematic coverage of the protocol. In this paper, we address this gap, and we propose an overview of the MAVLink protocol, the difference between its versions, and its potential in enabling Internet connectivity to unmanned systems.  We also discuss security aspects of MAVLink. To the best of our knowledge, this is the first technical survey and tutorial on the MAVLink protocol, which represents an important reference for unmanned systems users and developers.

\end{abstract}

%
\IEEEpeerreviewmaketitle

\section{Introduction}
Unmanned systems are autonomous platforms that can be easily programmed to perform missions with or without the intervention of a pilot. These systems can be aerial, also known as drones or unmanned aerial vehicles (UAVs), ground (UGV) or underwater. They typically communicate through wireless with a ground control station (GCS) that monitors their status and control their actions. UAVs embed special hardware and software called autopilot that controls the motion of the drone and monitors its status and used to communicate with GCSs using telemetry or WiFi communication. There are many autopilot software platforms for UAVs available including ArduPilot from 3DR \cite{team2016ardupilot}, Paparazzi UAV \cite{Paparazi} developed at Ecole Nationale de l'Aviation Civile (ENAC), Hangar autopilot \cite{hanger}, PX4 Flight Stack \cite{PX4}, MultiWii from Nintendo \cite{Multiwii} and several others. The most popular autopilot software is ArduPilot, which is an open-source project effectively maintained by a huge number of developers exceeding 400 contributors. It underpins different types of autonomous systems, including fixed-wing planes, (heli, tri, quad, hexa, and octo) copters, underwater vehicles and boats, and ground vehicles.  
In the heart of the ArduPilot system, Micro Aerial Vehicle Link (MAVLink) protocol is specified to ensure the communication between the unmanned systems and the ground stations. MAVLink is a well-established lightweight message serialization protocol specified for the unmanned vehicle systems, including drones. Lorenz Meier released MAVLink in 2009 under the LGPL license. MAVLink is designed as a Marshaling library, which means that it serializes messages of the states of the system and the commands that it has to execute into a specific binary format (i.e., a stream of bytes), that is platform-independent. The binary serialization nature of the MAVLink protocol makes it lightweight as it has minimal overhead as compared to other serialization techniques, (e.g., XML or JSON). The communication using MAVLink is bidirectional between the ground station and the unmanned system. Besides, given the binary serialization feature of MAVLink, its messages are typically of small sizes and can reliably be transmitted over different wireless mediums, including WiFi or even serial telemetry devices with low data rates. It also ensures the reliability and message integrity by a double checksum verification, in its packet header. All these features make the MAVLink protocol as the most popular among its peers for the communication between unmanned systems and GCS. 

Despite its popularity and the large community of users and developers, there is a significant lack of surveys about this protocol. New users/developers get usually confused due to the lack of structured references apart from some online documentation of some basic concepts. The only existing tutorial is entitled MAVLink Tutorial for Absolute Dummies (Part I), since 2013 \cite{balasubramanian2015mavlink}, (for which there is no part II!), which is rather an basic introduction to the elementary concepts of the MAVLink protocol. Also, in the last two recent years, there has been much research works around the MAVLink protocol. However, no survey discusses these works and classifies them. Thus, there is a desperate need in the community for a scientific research paper that provides both a tutorial and a survey on the MAVLink protocol to be the first reference for users, researchers and developers on this protocol and unveils the powerful features of this protocol and its main usages, extensions, and applications. 

This paper addresses this gap and provides a tutorial and a survey of the MAVLink protocol. The tutorial deals with presenting the main features of the MAVLink protocol and its two versions MAVLink 1.0 and MAVLink 2.0, and the most important messages specified in the protocol. In addition, we present the different tools and Application Program Interfaces (APIs) that developers need to parse and develop their control station programs that communicate with the unmanned systems. The survey part of the paper deals with presenting and discussing the main contributions proposed in the literature around MAVLink, which we classify in different categories including (\textit{i.}) enhancement and extension (\textit{ii.}) security, (\textit{iii.}) applications, (\textit{iv.}) integration with the Internet-of-Things (IoT), (\textit{v.}) Multi-UAV coordination and swarm. To the best our knowledge, this is the first and unique technical survey and tutorial that deals with the MAVLink protocol, which represents an indispensable reference for unmanned systems' users and developers. 

The remainder of the paper is organized as follows. Section II presents a general overview of the MAVLink protocol versions 1.0 and 2.0 and their header formats.  Section III discusses the security threats and vulnerabilities of MAVLink and presents the solutions proposed in the literature to address these security problems. Section IV provides a comprehensive state of the art that has contributed to the development of MAVLink and its applications in different contexts. Section V presents an overview of the software related to the MAVLink protocol, including ground stations, and simulation models. Section VI concludes the paper and discusses some future challenges.  

\section{The MAVLink Protocol} \label{sec:mav}

\subsection{Overview}

This section presents an overview of the MAVLink protocol, namely the transport and communication protocols supported in addition to the messages structures and serialization. 
In what follows, we use MAVLink messages and MAVLink packets interchangeably. 

\textbf{Note for readers:} The following subsections present low-level technical details about the protocol specification that are needed for practitioners and MAVLink developers. These details are summarized in tables whenever relevant.  

\subsection{Communication and Transport Protocols}
The MAVLink protocol defines the mechanism on the structure of messages and how to serialize them at the application layer. These messages are then forwarded to the lower layers (i.e., transport layer, physical layer) to be transmitted to the network. The advantage of the MAVLink protocol is that it supports different types of transport layers and mediums thanks to its lightweight structure. It can be transmitted through WiFi, Ethernet (i.e., TCP/IP Networks) or serial telemetry low bandwidth channels operating at sub-GHz frequencies, namely 433 MHz, 868 MHz or 915 MHz. The sub-GHz frequencies allow us to reach large communication ranges to control the unmanned system remotely. The maximum data rate can reach up 250 kbps, and the maximum range is typically expected to be 500 m, but highly dependent on the environment and level of noise and antenna setup. Table \ref{tab:telemetry-dev} presents the features of some commonly used telemetry devices. 

\begin{table*}[htbp]
\centering
\caption{An overview comparison between Ardupilot telemetry devices\cite{telmetry}}
\begin{tabular}{|c|c|c|c|c|c|}
\hline
\rowcolor[HTML]{EFEFEF} 
Telemetry device & Frequency        & Range              & Voltage    & Sensitivity          & RF transmit power \\ \hline
Bluetooth        &  Between 2402 and 2480 MHz, or 2400 and 2483.5 MHz           & 50 m                & 3.6 to 6 V & -80 dBm               & +4 dBm             \\ \hline
SiK Radio v2     & 900 MHz or 433 MHz & 500 m               & 3.3 V      & -121  dBm             & 20 dBm             \\ \hline
RFD900           & 900 MHz or 868 MHz & \textgreater{}40 km & 3.3 to 5 V & \textgreater{}121 dBm & +30 dBm            \\ \hline
Robsense         & 433 MHz           & 3-5 km             & 5 V         & -148  dBm              & 20 dBm             \\ \hline

\end{tabular}
\label{tab:telemetry-dev}

\end{table*}

The second alternative is to use a network interface, which is typically WiFi or Ethernet, and stream the MAVLink messages through IP Networks. In this case, the autopilot running the MAVLink protocol typically supports both UDP and also TCP connections at the transport layer between the ground station and the drone, depending on the reliability level required by the application. Of course, it is commonly known that UDP is a datagram protocol that requires no connection between the client and server and it has no mechanism to ensure that messages are reliably delivered, but provides a fast lighter weight alternative for real-time and loss-tolerant message streaming. On the other hand, TCP is a reliable connection-oriented protocol that provides better reliability of transfer thanks to its acknowledgment mechanism but could be subject to congestion and heavy management of the connection. The choice of the transport protocol is left to the user depending on the requirement he needs for the message exchange between the unmanned system and the ground station. 

\subsection{Message Types and Structures}
The unmanned system communicates with the ground station through the exchange of MAVLink messages, which are binary-serialized messages. \textit{Binary serialization} means that the content of the message is transformed into a sequence of bytes to be transmitted through the network. The receiver of the serialized message performs its deserialization in the opposite direction to reconstitute the original message sent.  This property of binary serialization has a significant benefit of reducing the size of the transmitted message to a maximum as compared to other types of serialization, such as XML or even the lighter weight JSON. Each MAVLink message contains a header appended to the message payload. The header carries out information about the message whereas the payload includes the data transported by the message. 

In the following section, we present the protocol headers of MAVLink 1.0 and that of the newer MAVLink 2.0 \cite{mavlinkv2} and the difference between each other. In the remaining of the paper, MAVLink refers to MAVLink 1.0 \cite{meier2013mavlink}, unless otherwise specified. 

\subsubsection{MAVLink 1.0 Protocol Header}

\begin{figure*}[htb]
\centering
\includegraphics[width=0.45\textwidth]{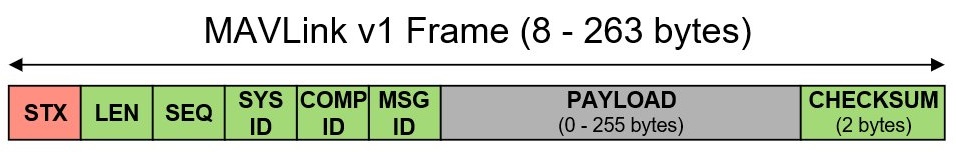}
\caption{The MAVLink 1.0 Header Structure}
\label{fig:first-msg}
\end{figure*}

As shown in Fig. \ref{fig:first-msg} there are eight important fields. 

The first field is STX and refers to the symbol that represents the start of a MAVLink frame. In MAVLink 1.0, STX is equal to special symbol \texttt{0XFE}. The second byte (LEN) represents the message length in bytes and is encoded into 1 Byte. 
The third byte (SEQ) denotes the sequence number of the message. It is encoded into 1 Byte and takes values from 0 to 255. Once it reaches 255, the sequence number is reset again to 0 and incremented in each generated message. The sequence number of message enabled to detect message losses in the receiver. The fourth byte SYS represents the System ID. Every unmanned system should have its System ID, in particular, if they are managed by one ground station. The System ID 255 is typically allocated for ground stations. One limitation of MAVLink 1.0 is that it restricts the number of drones managed by one ground station to 254 because the System ID is encoded in 1 Byte. 
The fifth byte is the Component ID, and it identifies the component of the system that is sending the message. There are 27 hardware types (i.e., components) in MAVLink 1.0. If there is no subsystem or component, then it is not used. The sixth byte represents the Message ID (MSGID), which refers to the type of the message embedded in the payload. For example, the message ID equal to 0 refers to a message of type HEARTBEAT, which indicates that the system is alive and is sent every one second. One more example with a Message ID equal to 33, which refers to a message that carries out the GPS coordinate of the unmanned system. The message ID is the essential information that allows to parse the payload and extract the information from it, based on the type of message. Each message contains a specified number of fields and serialized in binary format in a particular order, according to the standard specification.
The payload is located just after the message ID and can take a maximum of 255 bytes. Finally, the last two bytes are for the checksum. The CKA and CKB represent the Cyclic Redundancy Check (CRC) calculated with seed values A and B, respectively. The CRC ensures that the message has not been changed during its transmission and that both the sender and the receiver have the same message. It is calculated using the ITU X.25/SAE AS-4 hash of the bytes of the message, excluding the STX field (the hash is applied to 6+n+1 bytes, and the extra is the seed value). The seed is added at the end of the message when computing the CRC. 

The minimum message length of MAVLink 1.0 is 8 bytes for acknowledgment packets without the payload. On the other hand, the maximum length of a MAVLink 1.0 message is 263 bytes for full payload.

Summary and explanation of each MAVLink 1.0 header fields are presented in Table  \ref{tab:mavmsg}.

\begin{table*}[htb]
\centering
\caption{Explanation of MAVlink frame acronyms along with its contents.}
\label{tab:mavmsg}
\begin{tabular} {| p{1.5cm}|p{1cm}|p{12cm}|}\hline

\textbf{Acronym} & \textbf{Content} & \textbf{Description} \\ \hline
{STX} & 0XFE & It describes start of frame and will always be 0xFE as in official documentation of  MAVLink 1.0. \\ \hline
{LEN} & 0 to 255 & The value of LEN is described by length of payload. \\ \hline
{SEQ} & 0 to 255 & The sequence of packet is shown in this part of message. Such as 0 represent the first message. It is used for detecting lost MAVlink packets. \\ \hline
{SYS} & 1 to 255 & This field represents the ID of the unmanned system. \\ \hline
{COMP} & 0 to 255 & This field represents which component in the system is sending the message. \\ \hline
{MSG} & 0 to 255 &This field represents the message type.\\ \hline
{Payload} & 0 to 255 bytes & This carries out the real data of the message, which depends on the message type.\\ \hline
{CKA and CKB (CRC) or checksum} & Two bytes contents &The CKA and CKB is referred as checksum. The signing of packet happens from Least Significant Bit (LSB) to Most Significant Bit(MSB). \\ \hline
\end{tabular}

\end{table*}


\subsubsection{MAVLink 2.0 Protocol Header}
\begin{figure*}[htb]
\centering
\includegraphics[width=0.8\textwidth]{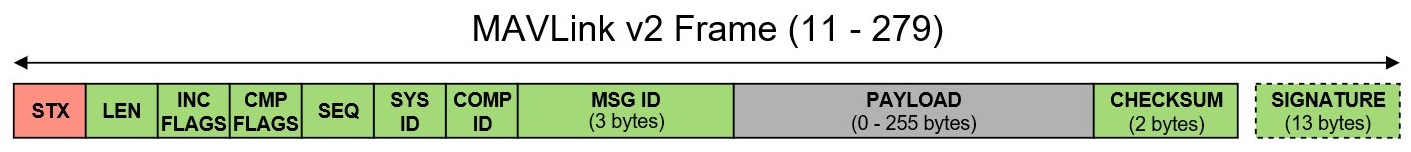}
\caption{MAVLink 2.0 Header}
\label{mavlink2header}
\end{figure*}

The MAVLink 2.0 protocol header was released in early 2017 and is the current recommended version. It is backward compatible with the MAVLink 1.0 version and includes several improvements over the MAVLink 1.0 version. We first start with presenting the MAVLink 2.0 protocol header; then we highlight the main differences between the two versions. Fig. \ref{mavlink2header} shows the header structure of MAVLink 2.0.

The MAVLink 2.0 header shares all the fields with MAVLink 1.0 header, and adds its new fields, in addition to the changing the size of some existing fields. 
The first byte is the start-of-text marker, and its specific value is \texttt{0xFD} for MAVLink 2.0 (as opposed to \texttt{0xFE} for MAVLink 1.0). Thus, the parser has to first recognize this character before being able to parse the remaining fields of the MAVLink 2.0 message. The payload length is the next field and is the same as in the legacy protocol. MAVLink introduces two new flags before the sequence number (SEQ) of the message. 
The first flags are \textit{Incompatibility flags}, which are flags that affect the message structure. The flags indicate whether the packet contains some features that must be considered when parsing the packet. For example, an  Incompatibility flag equal to $0x01$ means that the packet is signed and that a signature is appended at the end of the packet. The second flag is \textit{compatibility flags}, which does not affect the structure of the message. It indicates flags that can be ignored if not understood and it does not prevent the parser from processing the message even if the flag cannot be interpreted. For example, this may refer to flags that indicate the priority of the packet (e.g., High Priority) as it does not affect the packet structure. The sequence number (SEQ), the System ID (SYSID) and Component ID (COMPID) are the same as in the MAVLink 1.0 protocol header. However, the Message ID (MSGID) is encoded into 24 bits instead of 8 bits in the previous version, which allows a much higher number of message types in MAVLink 2.0, reaching up to 16777215 possible types. It is not clear what is the reason to design such a huge space of possible message types, as the number of possibilities is overly large.
The Payload field can take up to 255 bytes of data, which depends on the message type. The checksum is similar to its peer in MAVLink 1.0. Finally, MAVLink 2.0 uses an \textit{optional} Signature field of 13 bytes to ensure that the link is tamper-proof. This features significantly improve security aspects of the MAVLink 1.0 as it allows the authentication of the message and verifies that it originates from a trusted source. The signature of the message is appended if the incompatibility flags are set to $0x01$. 

The 13 bytes of the message signature contain the following fields:
\begin{itemize}
    \item \textit{LinkID}: it is one byte that represents the ID of the link (channel) used to send the packet. A link or a channel can be WiFi or telemetry and can be combined. Every channel used to send data should have its own LinkID. It provides a means for multi-channel unmanned system control using MAVLink 2.0.    
    \item \textit{timestamp}: it is encoded with 6 bytes in 10-microsecond units since 1st January 2015 GMT. It increases for every message sent over the channel. It is applied to every stream where a stream is defined by the tuple \texttt{(SystemID, ComponentID, LinkID)}. The timestamp is used to avoid replay attacks.
    \item \textit{signature}: it is encoded in 6 bytes for the message and is calculated based on the complete message, timestamp, and the secret key. The signature includes the first 6 bytes (48 bites) of the SHA-256 hash applied to the MAVLink 2.0 message (excluding the signature, and including the timestamp). The secret key is a shared symmetric key of 32 bytes stored on both ends, namely the autopilot, and the ground station or the MAVLink API. 
\end{itemize}

The signature of the MAVLink 2.0 message has consequences on how to process incoming MAVLink messages. If the message is signed, then it is discarded if (\textit{i.}) the timestamp of the received message is older than the previous packet received from the same stream identified by \texttt{(SystemID, ComponentID, LinkID)} tuple, (\textit{ii.}) the computed signature at the reception is different from the signature appended to the message. This may infer a data alteration in the message, (\textit{iii.}) the timestamp exceeded one minute as compared to the local system's timestamp.  
If the message is not signed, then the acceptance/rejection of the packet is implementation specific.

\subsection{MAVLink Messages Types}

\begin{figure}[]
\centering
\includegraphics[width=0.5\textwidth]{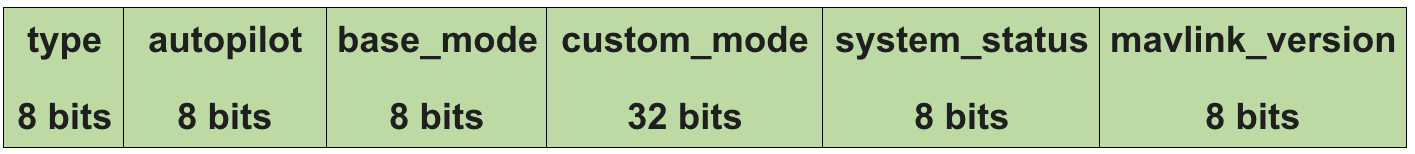}
\caption{MAVLink Heartbeat Message}
\label{fig:mav-heart-msg}
\end{figure}

MAVLink defines several types of messages, which are identified by their Message ID. Messages with Message IDs lower than 255 are common for both MAVLink 1.0 and MAVLink 2.0, and those with Message IDs higher than 255 are specific to MAVLink 2.0. As mentioned in the previous section, the Message ID in MAVLink 1.0 is encoded in only 8 bits and was extended to 24 bits in MAVLink 2.0. 

We categorize MAVLink messages into two classes:
\begin{itemize}

    \item \textit{State messages}: these messages are sent from the unmanned system to the ground station and contain information about the state of the system, such as its ID, location, velocity, and altitude. 
    
     \item \textit{Command messages}: they are by the ground station (or user program) to the unmanned system to execute some actions or missions by the autopilot. For example, the ground station can send a command to a drone to take off or to land or to go to a waypoint or even a to execute a mission with several waypoints.

\end{itemize}

Considering the large number of MAVLink messages, the comprehensive description of all these messages is out of the scope of this paper and can be found in details in \cite{mavlinkspec}. In what follows, we present the most relevant state and command messages used in common implementations of autopilots. More messages are presented in Table \ref{tab:mav1}.
\\
\subsubsection{State Messages}

There are several types of state messages defined in MAVLink. 

\par

\textbf{HEARTBEAT message}: The HEARTBEAT message is the most important message in MAVLink, and its structure is depicted in Fig. \ref{fig:mav-heart-msg}. It indicates that the vehicle system is present and active. The unmanned system periodically sends the heartbeat message  (in general every second) to the ground station to inform the GCS that it is active. The heartbeat is a required message. In addition to the header, the message payload contains essential information about the unmanned system. The first field is the \texttt{type}, which indicates the type of the  Micro Aerial Vehicle. According to the latest specification in \cite{mavlinkspec}, there are 33 pre-defined types defined in the \texttt{MAV\_TYPE}, including quadrotor (\texttt{MAV\_TYPE\_QUADROTOR = 2}), helicopter (\texttt{MAV\_TYPE\_HELICOPTER = 4}), fixed wing (\texttt{MAV\_TYPE\_FIXED\_WING = 1}), and several others. The autopilot field indicates the type of autopilot. There are several types defined in the \texttt{MAV\_AUTOPILOT} enumeration structure. For example, \texttt{MAV\_AUTOPILOT\_GENERIC = 0} indicates a generic autopilot, \texttt{MAV\_AUTOPILOT\_ARDUPI\-LOTMEGA = 3} indicates ArduPilot autopilot, \texttt{MAV\_AUTOPI\-LOT\_PX4 = 12} for PX4 autopilots.

    \par The \texttt{base\_mode} field indicates different operation modes. Understanding the \texttt{base\_mode} is crucially important to correctly parse the heartbeat message and extract useful information out of it. It is encoded in 8 bits. There are 8 pre-defined flags, from $2^0$ to $2^7$.  Here are the eight different modes:
    
    \begin{itemize}
        \item \texttt{Flag = 1} is reserved for future use
        \item \texttt{Flag = 2} means that the test mode is enabled. This mode is used for temporary tests and not used for regular flights. 
        \item \texttt{Flag = 4} means that the autonomous mode (AUTO) is enabled. This means that the unmanned system is operating autonomously by navigating to the goal waypoints sent to it from the ground station. In AUTO mode, a mission is loaded to the autopilot. A mission consists of a set of \textit{several} waypoints that the system has to navigate.  
        \item \texttt{Flag = 8} means that the GUIDED mode is enabled. In GUIDED mode, a mission consists of a \textit{single} waypoint sent to the system. The system then navigates to the specified location autonomously. 
        \item \texttt{Flag = 16} means that the system stabilizes its attitude (orientation and altitude), and possibly its position, by automatic control. This requires external sensors like GPS in an indoor environment, altitude sensors (barometer, LIDAR) or motion capture for indoor positioning to be able to hover in a stable state. The system needs external control inputs to make it move around. 
        \item \texttt{Flag = 32} means that the hardware in the loop simulator is activated, i.e., all motors and actuators of the motors are blocked while the internal autopilot is fully operational. 
        \item \texttt{Flag = 64} means that manual mode is enabled, which requires that the pilot manually control the system using a remote control input. In manual control, there is no automatic control made by the autopilot. 
        \item \texttt{Flag = 128} the system is in \textit{armed} state, which means that motors are enabled/running and can start the fly. 
    \end{itemize}
    
 The custom mode is also essential. It indicates autopilot specific flags that are interpreted in addition to the base mode. It is used in heartbeat message parsing to determine the flight modes of the autopilot system. There are pre-defined values for the custom mode including 0 for manual flight mode, 4 for guided mode, 10 for auto mode, 11 for RTL mode, 9 for LAND mode, 2 for ALT\_HOLD, and 5 for LOITER. In the next section, we provide a comprehensive overview of the different flight modes in ArduPilot systems supported by MAVLink. 
 
 The \texttt{system\_status} field represents a flag that indicates the system state. There are night states defined as of the latest specification \cite{mavlinkspec}:

 \begin{itemize}
        \item \texttt{system\_status = 0} refers to a system that is not initialized system or an unknown state. 
        \item \texttt{system\_status = 1} indicates that the system is booting. 
        \item \texttt{system\_status = 2} means that the system is performing a a calibration. In fact, the sensor calibration is a very important phase to make sure that flight sensors such as Inertial Measurement Units (IMU), and Compasses are consistent and run as expected.  
        \item \texttt{system\_status = 3} it means that the system is in \textit{standby} mode and can be started at any time. 
        \item \texttt{system\_status = 4} indicates that the motors are engaged and that the system is \textit{active} and might be airborne. 
        \item \texttt{system\_status = 5} indicate potential errors and that the system is in \textit{critical} state, although it can still navigate. This can happen for example during temporary interference or battery level starting to be low, etc. 
        \item \texttt{system\_status = 6}  this means an \textit{emergency} situation where the unmanned system lost control over some parts and is in distress situation. The system may have already been crashed. 
        \item \texttt{system\_status = 7}  indicates that the system has started its power-off process and is now \textit{shutting down}.  
        \item \texttt{system\_status = 8}  indicates that the system is \textit{terminating} itself and ending its flight.  
        
    \end{itemize}
 
 Finally, \texttt{mavlink\_version} field indicates the MAVLink version. It is not editable by the user and is set by the protocol.



\par

\textbf{System Status message:}  The system status message has a Message ID equal to 1 and is composed of data about the onboard control sensors embedded into the unmanned system and specifies which of these sensors are enabled/disabled and which sensors are operating or having errors. It also carries out data about the battery status and the remaining voltage, which is useful to track the battery level of the unmanned system. Besides, it provides information about the communication errors and the ratio of dropped packets in the communication link. The information about battery and communication link are crucial to take appropriate failsafe action when the battery level goes down, or the communication quality deteriorates. In this case, the unmanned system can be pre-programmed to execute a failsafe operation in case of low battery level or bad communication quality such as landing and going back home for an unmanned aerial system. 

\par 

\begin{figure}[htb]
\centering
\includegraphics[width=0.5\textwidth]{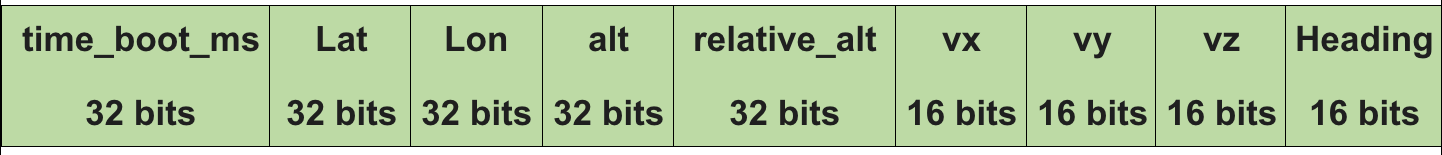}
\caption{System status message}
\label{fig:mav-pos-msg}
\end{figure}

\textbf{Global Position message:} The global positioning message has an id equal to 33 and represents the filtered GPS coordinate provided by the Global Positioning sensor. It is illustrated in Fig. \ref{fig:mav-pos-msg}. This message carries out important information of the unmanned system related to its global positions, namely, the GPS latitude (\texttt{lat}), longitude (\texttt{lon}) and also \textit{absolute} altitude (\texttt{alt}). These three values are encoded into four bytes (32 bits). The values of (\texttt{lat}) and (\texttt{lon}) must be divided by $10^7$, to get the real floating GPS value, it is needed to divide them by $10^7$. The altitude is expressed in millimeters. The message also contains a \textit{relative} altitude field (\texttt{relative\_alt}), which represents the altitude relative to the takeoff ground point of the unmanned aerial system. It is different from the absolute altitude. For example, the absolute altitude of Riyadh city is 612 meters, which corresponds to a relative altitude of 0 meters, as the drone is on the ground. If the drone takes off at a relative altitude from the ground of 10 meters, then its absolute altitude becomes 622 meters. Besides, this message also carries out information about the linear speed of the unmanned system along the 3 axis \texttt((x,y,z) in addition to orientation referred to as heading. This information is collected from the GPS sensor, and it can also be read from other sensors such as the Inertial Measurement Unit (IMU) or the compass, which are available in other MAVLink messages. 
    
\subsubsection{Command Messages}

There are several command messages in MAVLink that give the ability to request the unmanned system to perform certain actions. In what follows, we provide an overview of the most important commands. Table  \ref{cmd_table} presents a summary of a selected set of MAVLink commands.

\par 

\begin{figure}[htb]
\centering
\includegraphics[width=0.5\textwidth]{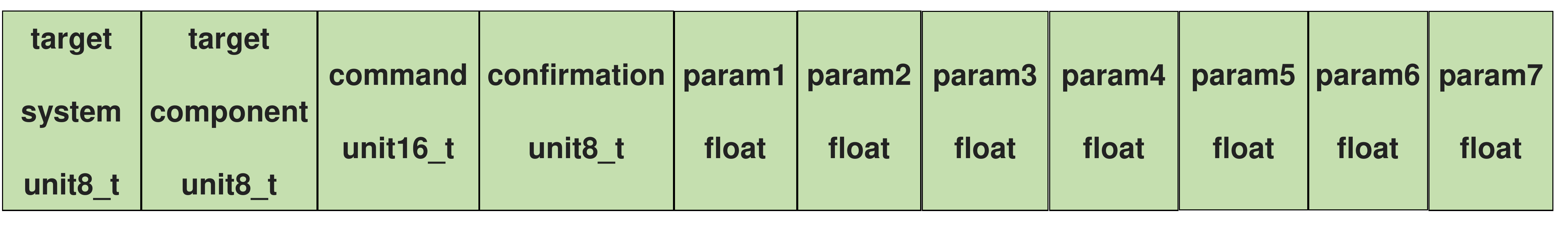}
\caption{Command long}
\label{fig:mav-command-long-msg}
\end{figure}

\textbf{COMMAND\_LONG:} The COMMAND\_LONG is a multi-purpose command that allows sending different types of commands depending on the command type of the message and its parameters.  The COMMAND\_LONG message has a Message ID equal to 76 and is defined with 11 fields, as illustrated in Fig. \ref{fig:mav-command-long-msg}. The \texttt{target\_system} field and the \texttt{target\_component} field specify the system that will execute the command, and its underlying component. 

The command field refers to the type of command to be executed. It is defined in the \texttt{MAV\_CMD} command enumerations.  Also, for each command, a set of parameters relevant to the command can be set. In what follows, we provide a summary of some commands and their respective parameters. 

\begin{table}[!htbp]
\caption{Selected List of Important MAVLink Commands}
\label{cmd_table}
\begin{tabular} {| p{1.3cm}|p{1cm}|p{1.8cm}|p{2.5cm}|}\hline

\hline
Command             & Command ID & Parameters      & Description                                                                                               \\ \hline
TAKEOFF             & 22         & param7: double  & this command makes the aerial unmanned system takeoff at an altitude specified in param1                  \\ \hline
LAND                & 21         & No parameters   & this command makes the aerial unmanned system land to ground                                              \\ \hline
GET\_HOME & 410        & No parameters   & this command allows to get the Home position, which is the first waypoint in the mission list             \\ \hline
SET\_HOME & 179        & \begin{tabular}[c]{@{}l@{}}param5: double\\ param6: double\\ param7: double\end{tabular}   & this command allows to get the Home position, which is the first waypoint in the mission list             \\ \hline
ARM  DISARM          & 400        & param1: boolean & this command allows to arm the motors if param1 is set to true, and disarm them if param1 is set to false \\ \hline
\end{tabular}

\end{table}

For example, the command ID with 21 refers to the LAND command, and it has no parameter. Some other commands have parameters like the takeoff command (ID=22), which has to specify the takeoff altitude in \texttt{param 7}, and the command arm/disarm which specifies a Boolean value in \texttt{param1} to indicate whether to arm or disarm the motors. There are around 60 commands types defined in the MAV\_CMD command enumerations. The confirmation field indicates if the message was sent for the first time with value 0, and other values represent a confirmation of the message. The 7 parameters depend on the type of command. For example, for the LAND command, all seven parameters are useless. In the takeoff command, the seventh parameter represents the altitude requested for the takeoff. 

\textbf{The Mission Item Command:} The mission item command has a Message ID equal to 39 and allows to send a waypoint to an unmanned system so that it navigates autonomously to that specific waypoint in GUIDED mode. Every mission item command message has a sequence number that specifies its number in the mission, starting from 0, which specifies the home location. It also has three fields \texttt{(x,y,z)}, which specify the coordinates of the waypoint. However, the coordinate must be specified with respect to a reference frame. Thus, the message has a field called \texttt{frame}, which specifies the reference coordinate frame of the waypoint. This parameter is important because it is essential to interpret the meaning of the coordinates \texttt{(x,y,z)}. For example, if we set the waypoint coordinate as (24.68773, 46.72185, 10) with reference to the global frame MAV\_FRAME\_GLOBAL, this would mean to go to the waypoint at GPS location  (24.68773, 46.72185) and an \textit{absolute} altitude of 10 meters with respect to sea level. For example, in Riyadh city (Saudi Arabia), the absolute altitude is 620 m, so this would be that the drone may go down and crash to the ground because the target altitude is lower than that of the ground. 
However, usually, we want to specify the location with respect to the ground (e.g., 10 meters above ground) and for this, we need to specify the reference frame to be a global frame but with \textit{relative altitude} namely MAV\_FRAME\_GLOBAL\_RELATIVE\_ALT. The command also specifies the target system and target component as in other command messages.



\begin{table}[!htbp]
\caption{Selected List of Important MAVLink Messages}
\label{tab:mav1}
\begin{tabular} {| p{0.9cm}|p{2cm}|p{4.8cm}|}\hline
\textbf{Message Type} & \textbf{Message Representation} & \textbf{Description} \\ \hline

{0} & HEARTBEAT & It is the most important message in MAVLink that tells if the unmanned system is alive or not. \\ \hline
{1} & SYS\_STATUS & It defines the unmanned system state including onboard sensors, communication quality and battery status. 
\\ \hline
{2} & SYSTEM\_TIME & It defines the system time of the master clock that is usually the onboard clock. \\ \hline
{5 and 6} & CHANGE OPERATOR (CONTROL, ACK) & They represent the request to take control over the unmanned system and its corresponding acknowledgment \\ \hline
{20, 21, 22, 23} & PARAM REQUEST (READ,  LIST),PARAM (VALUE ,SET) & These four important messages are related to on-board parameters whose value can be obtained by GCS or can be set by GCS. For example, it is possible to request to read the SystemID parameter or to change it. \\ \hline
{24, 25, 33, 48, 49, 123, 124, 127} &GPS RAW INT, GPS STATUS, GLOBAL POSITION INT, SET\_GPS GLOBAL ORIGIN , GPS GLOBAL ORIGIN, GPS INJECT DATA, GPS2 RAW, GPS RTK, & 
These messages are related to the GPS sensor information, such the raw GPS value, the Global Position value, etc.  
\\ \hline
{26,27} & SCALED IMU and RAW IMU & These messages contain the scaled and raw IMU sensor data according to 9 degrees-of-freedom including acceleration, gyro (angular speed) and magnetic field all in three axes\\ \hline
{37 up to 47 and 51} & MISSION related messages & There are 10 messages defined for missions request a mission, or set a waypoint in a mission, clearing a mission (i.e., delete all its waypoints), or getting acknowledgement for a mission, etc. A mission is defined as a set of waypoints sent to the unmanned system to navigate to them in the autonomous mode. 
\\ \hline
{34, 35, 50, 65, 70, 92} & RC\_CHANNELS  SCALED, RAW, PARAM MAP RC, OVERRIDE, HIL RC INPUTS RAW & All these messages are for RC control to get the channels, write data or controls.  \\ \hline
{75, 76} & COMMAND (INT, LONG, ACK) &It sends the command to the unmanned system for performing actions such  as (a) navigation commands (b) do commands (start, jump etc.) or (3)   condition commands,  all listed in   \textcolor{black}{MAV CMD.}\\ \hline
\end{tabular}%

\end{table}

\subsection{Flight Modes}

Understanding flight modes is crucially important to be able to pilot a unmanned system running Ardupilot and the MAVLink protocol. There are several flight modes that were defined by Ardupilot. In this section, we present the most important and more common flight modes.

\begin{itemize}
    \item \textbf{The STABILIZE mode} 
    In this mode, the unmanned system is fully controlled by the user, position, altitude, and heading. The unmanned system will respond to every input from the RC controller controlled by the user, and it is up to the user to compensate any drift made by the unmanned system. 
    
    It is recommended that users immediately switch to the STABILIZE mode to manually control the unmanned system, when the autopilot fails to control the vehicle in any other autonomous mode. 
    
    It has to be noted that it is possible to download dataflash log files from your unmanned system, to analyze the flight performance by opening it in mission planner in the graph tab.
    
    

    \item \textbf{ALTITUDE HOLD}: A more comfortable mode to control the unmanned system is the ALTITUDE HOLD or ALT HOLD mode where the user does not have to worry about maintaining a fixed altitude for the unmanned system, as the autopilot will take care of controlling automatically using a PID controller the altitude. The user will have to take care of controlling the direction and position of the unmanned system manually. 
    
\begin{figure*}[thb]
\centering
\includegraphics[width=1\textwidth]{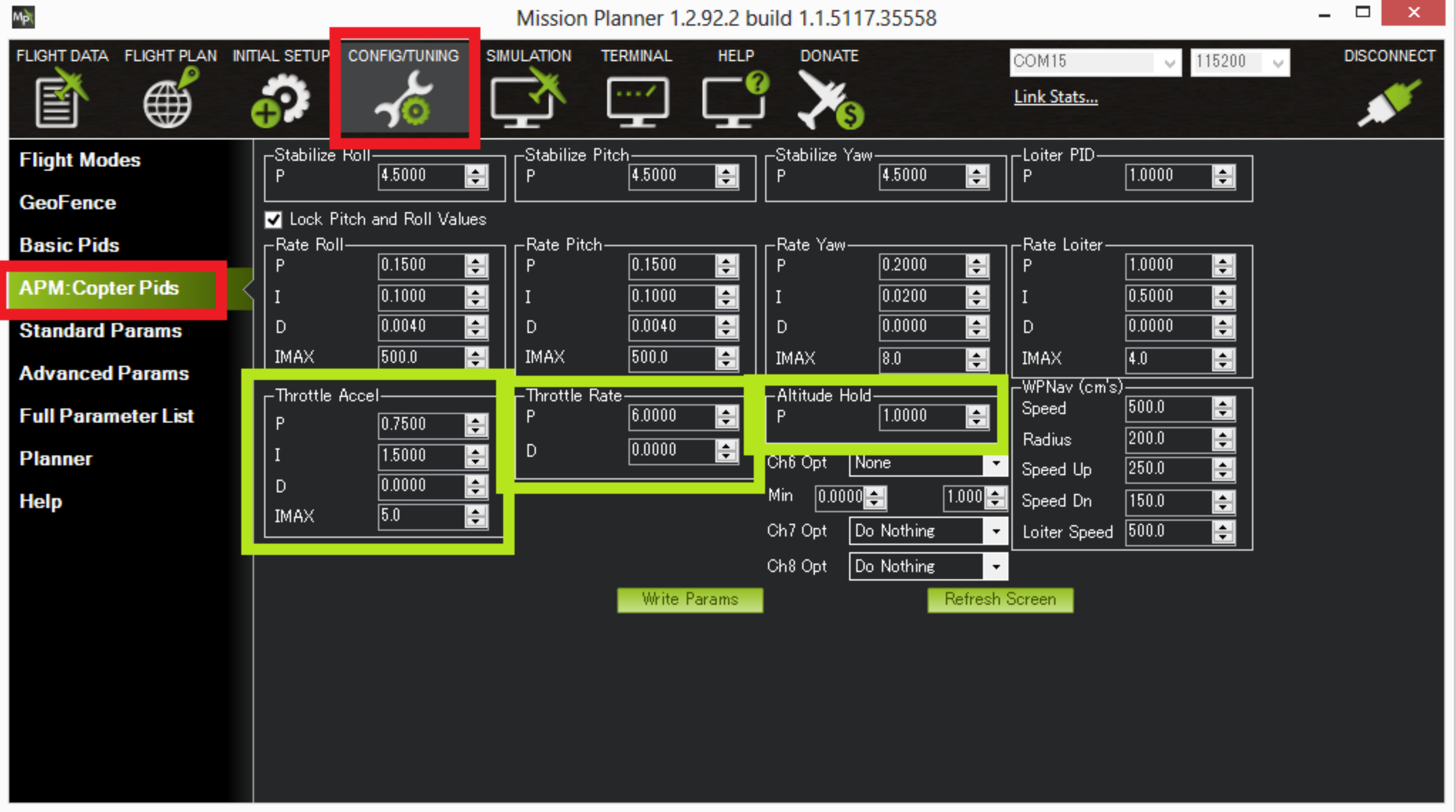}
\caption{Settings of the PID controller }
\label{fig:pid}
\end{figure*}

    The ALT\_HOLD mode automatically controls the altitude of the unmanned system by the autopilot. However, it does not control the heading, and position, which are left to the user. This mode is more recommended for newbies than the STABILIZE mode and does not require a GPS because it estimates the altitude with the barometer. 
    
    It is possible to tune the setting of the PID controller of the ALT\_HOLD mode in APM mission Planner as illustrated in Fig. \ref{fig:pid}. The altitude is maintained with proportional controller that estimate the error between the desired altitude and the actual altitude and tune the vertical acceleration proportionally to that error. The proportional gain can be set through a ground station as illustrated in Fig. \ref{fig:pid}. The Proportional gain must be carefully set because a very high gain will make the control more aggressive and less stable, whereas a very low gain will make the control very slow and non-responsive.

    
    

    \item \textbf{LOITER}: An even more accessible mode to control the unmanned system is the LOITER mode, which maintains the current location, orientation, and altitude of the unmanned system once the user does not provide input to the RC controller. This mode is similar to the STABILIZE mode, but the unmanned system will control its position, heading, and altitude once the user gets his fingers out of the RC sticks. The LOITER mode requires a GPS 3D fix to work with, or optical flow. 
    
    
    It has to be noted that it is not possible to arm the vehicle in LOITER mode only if (1) GPS has 3D Lock, (2) HDOP is smaller than 2.0. As such, this mode requires a GPS 3D fix to work with. 
    
    To achieve excellent LOITER performance, it is essential to have (1) GPS Lock, (2) low magnetic interference of the compass, and (3) low vibration.
    
    The PID controller gains can be tuned from the Mission Planner ground station or similar ground stations. 
    
    The LOITER SPEED represents the max horizontal speed in cm/s and is typically equal to 500 cm/s (which is equivalent to 5m/s). The default configuration is that the maximum acceleration is equal to the half of the LOITER SPEED (i.e., $2.5m/s^2$).
    
    The parameters of the LOITER mode can be configured through a ground station as illustrated in Fig.6 by setting PID control gains of the altitude, position and orientation (Yaw, Pitch and Roll). 
    

    \item \textbf{LAND}: The LAND mode will force the unmanned system to land to the ground.
    
    \item \textbf{RTL}: The RTL mode, also called Return to Launch, will force the unmanned system to return to start position where it performed the TAKEOFF. 
    
    LAND and RTL mode are used in case of violation of navigation safety and geofence, for example, it is possible to program on the autopilot that if the battery goes under a certain level, then the unmanned system needs to LAND immediately or return to start position automatically. This is called GEOFENCE.

    \item \textbf{GUIDE}: The GUIDED mode is essential and operates only with GPS mode. When the GPS of the unmanned system performs 3D fix and is activated, then the unmanned system can be sent to navigate autonomously to a particular GPS coordinate through the ground station. It is called GUIDED mode because the unmanned system is guided by the user to navigate autonomously to a specific waypoint chosen by the user. 
    
    The GUIDED mode is used in conjunction with GPS and allows the user to send the unmanned system to specified waypoint defined by their GPS coordinates. It is not possible to arm the unmanned system in GUIDED mode only when the GPS has the 3D fix.
    
    In GUIDED mode usually, a ground station is used to send a navigation waypoint to the unmanned system to navigate to it. It is therefore important to have a telemetry device connected to the unmanned system and the ground station to allow long-range communication between them. 
    
    Typically, a user needs to click on a point on the map, in GUIDED mode and the unmanned system will plan its path and move towards the goal.

    \item \textbf{AUTO}: The AUTO mode refers to the autonomous mode, where the unmanned system will follow a predefined mission. A Mission is a set of waypoints stored in the unmanned system autopilot. When the AUTO mode is selected, the unmanned system will autonomously go to each waypoint in the same order as they are stored.

\end{itemize}


\section{Security Issues of MAVLink}

\subsection{MAVLink Security Requirements }
Many research works dealt with the security of unmanned systems but much less addressed MAVLink security issues. Among them, several works tackled the security requirements that must be taken into account in the MAVLink protocol. In what follows, we summarize the most important requirements in terms of confidentiality, integrity, availability, authentication, non-repudiation, authorization, and privacy in order to secure MAVLink communications.
\begin{itemize}
    \item \textbf{Confidentiality:} It is an essential requirement to guarantee private data exchange between entities. It protects against attacks pertaining to non authorized disclosure of secret information exchanged in the network between the unmanned system and the ground station, since an adversary might be able to intercept commands sent from GCS to the unmanned system, or steal other system state data transmitted from the GCS to the unmanned system.

    \item \textbf{Integrity:} It is mandatory to secure communications between GCSs and UAVs by ensuring the integrity of exchanged data. Integrity is required to ensure that telemetry information sent from the unmanned systems and control signals sent from the GCS have not been intentionally or unintentionally interrupted, altered or modified. It protects against threats that pertain to unauthorized information modification.
    
    \item \textbf{Availability:} 
  The communication between UAVs and GCSs should be available, as well as the information itself is always available when needed or requested, even if a fault has occurred on the UAV system or an attacker tries to jam the UAV/GCS channel.
  
   It is fundamental that all the elements of the UAV system are operating and performing their requested functions and expected services when needed. Considering system maintenance, it is mandatory to ensure the continuous operation of the service without interruption, and to guarantee that the system performance is maintained so that the system keeps its availability uninterrupted during operation.
Besides, for the accessibility to the UAV, the service must be available when the user needs it. 
    
    \item \textbf{Authentication:} In a network of unmanned systems, multiple entities are participating and exchanging information in the network. Authenticating these entities and information origins is mandatory.
Authentication allows each node to verify the origin of the data transmitted, i.e., make sure that the message is effectively received from an authentic source. The authentication of the unmanned system by the GCS is highly critical to make sure the GCS is controlling an authorized drone, not a fake one. Also, it is essential that GCS can also be authenticated so that an unmanned system does not send its state or accept commands from a hacked/fake ground station. Therefore, it is mandatory to authenticate both ends to ensure that data sources are trustworthy.

    \item \textbf{Non-repudiation:} Any entity in UAV network does not deny that it has sent or received data or control commands. 

    
    \item \textbf{Authorization:} It refers to the ability of a system to permit access to information, and which actions these entities are allowed to perform. In the case of MAVLink only authorized GCSs and UAVs, can enter the network, are permitted to exchange telemetry data and send control command.
    
    \item \textbf{Privacy:} Exchanged information between GCSs and UAVs includes sensitive information about location, battery status, speed, weather, wind speed, mission status, etc. This private information must not be leaked to unauthorized third parties. Thus, it is mandatory to preserve the privacy of communications, hide UAVs and GCSs' identities and protect sensitive information issued by UAVs and GCSs from intruders.
\end{itemize}

\subsection{MAVLink Security Threats }

Communication between UAVs and GCS is established by a communication protocol via a wireless channel, which makes them vulnerable to various attacks since the communication protocol MAVLink does not support security procedures. Both confidentiality and authentication mechanisms are not natively supported. The GCS exchange data with UAVs through an unauthenticated channel and without encryption. These connections can be easily hacked, someone with an appropriate transmitter can communicate with the drone, inject commands into an existing session and easily launch attacks on UAVs.

The open nature of communications makes MAVLink vulnerable against various malicious attacks. These attacks can be classified as Interception (Attacks that compromise data confidentiality), Modification (Attacks that compromise data integrity), Interruption (Attacks that compromise data availability) and Fabrication (Attacks on authenticity). In the following, we detail further these threats:
\subsubsection{Confidentiality and privacy attacks}

 In this category of attacks, an intruder gets an unauthorized access to confidential and sensitive information by intercepting data, commands or messages exchanged between UAV and GCS. The confidentiality and privacy of the information are affected in this category of attack. Such attacks concern eavesdropping, identity spoofing traffic analysis and unauthorized access and are a result of deficiencies in the security of MAVLink.

\begin{itemize}
    \item \textbf{Eavesdropping (Communication capture):} Due to the lack of encrypted connections, an attacker listens to the communication happening between UAVs and GCS, eavesdrops on the information exchanged between UAVs and GCS directly from the open environment. This kind of attacks is exploited by the adversary to obtain information about the UAV, and consequently perform more elaborated attacks (active attacks).
The attacker captures control data and commands sent from the GCS to the UAV, to be used in a replay or a fabrication attack. Telemetry data broadcasted from UAVs to the GCS is intercepted by an adversary to gain knowledge about the UAVs location and flying speed. Eavesdropping is a passive attack that breaches the confidentiality and privacy of the control signal and telemetry data. The lack of data encryption and authentication in communication stimulates such attacks.

\item \textbf{Identity spoofing:} The MAVLink Communication protocol is unencrypted and uses the System IDs to identify the drone which sends or is expected to receive the messages. System ID is sent in clear within the unencrypted MAVLink header. Thus, an attacker can compromise the communication link to get the identity of the sending system.

\item \textbf{Traffic analysis:} Traffic analysis is a passive attack. An intruder may collect exchanged data to infer specific data to reveal specific patterns about the  communication between UAV and GCS. It can be any useful information such as frequency of MAVLink communication, size of MAVLink packets, etc. Traffic analysis is a method to gather useful and sensitive information that potentially can be used in other attacks. 
\item \textbf{Unauthorized access:} It occurs when the attacker gets access to the UAV and/or GCS, their services and resources using duplicated SYSID or COMPID. This attack usually results in unauthorized disclosure of GCS and telemetry information from UAV.
\end{itemize}

\subsubsection{Integrity attacks}
 The integrity of MAVLink can be compromised by modifying the data being sent. Violation of the MAVLink integrity allows the following attacks:

\begin{itemize}
    \item \textbf{Man-in-the-middle:} In the MAVLink communication protocol, messages are sent in plain text, which represents vulnerability and a threat to network security. The M-I-T-M attack can be successfully established in the channel. The attacker is located between UAVs and GCS and listen to the exchanged communications. The attacker can infer the content of the intercepted MAVLink payload and reconstruct commands. He can replay previously recorded packets, modify control and telemetry data and send these wrong data back to the GCS or the UAV. Thus, the integrity of the control data and telemetry data is hampered.
    \item \textbf{Hijacking (Unauthorized Command Injection):} There is a possibility, whenever a M-I-T-M attack against a UAV is successful, the attacker sends unauthorized commands to the UAV to takes its control from its GCS while allowing the GCS to believe that it is still controlling the UAV. Once the drone is under his control, the attacker can catch and withhold the UAV. There are two ways to hijack a drone using MAVLink vulnerabilities:
    (\textit{i.})\texttt{(Skyjacking):} Exploiting MAVLink's lack of authentication, a drone called Skyjack hacks other drones by using airplay-ng software to force disconnecting the authentic user from the drone by injecting de-authentication messages. Since the
drone does not authenticate users, an hacker may easily connect to it and take control over the device as soon as the WIFI connection is established.
(\textit{ii.})\texttt{(Radio Jacking):} It is another way to hack a UAV using the MAVLink's vulnerability. To control the drone using telemetry via MAVLink, it is mandatory to set up the NetID to connect to the drone. If an intruder recognizes the NetID field, he can easily hijack the UAV through the use of an antenna with the sniffed NetID to transmit malicious MAVLink packets and false information.
     \item \textbf{Replay attack:} Due to the ease of message capture and open nature of communications, the MAVLink protocol is vulnerable to replay attacks. A malicious user records the control data sent to the drone and replays them later to misuse the drone and produces an unauthorized effect. This attack may cause loss of control over the drone and possible crash.
    \item \textbf{Message modification:} Modification of messages means altering the contents of the data packet. The attacker captures the control data sent by the GCS, modifies them and sends wrong data back to the UAV. As a result, the GCS's control data are misinterpreted by the drone resulting in the drone being uncontrollable. The integrity of the control data and telemetry data is compromised.
    \item \textbf{False location update:} An attacker can send spoofed messages to the GCS using the data link that seems to be from a UAV containing false UAV location data using Scapy (a packet manipulation tool) to spoof heartbeat messages. This attack makes GCS believing that the UAV is in another location, or is following a wrong trajectory.

\end{itemize}
\subsubsection{Availability attacks} Attacks that compromise the availability of MAVLink can be achieved through interruption of the link used to exchange data between the drone and the ground station. There are several means on how to perform this attack, in particular through jamming, deletion attack, falsifying signals and Denial of Service (DoS/DDoS) attacks.

\begin{itemize}
\item \textbf{Jamming:} This attack affects system availability.
An attacker who is trying to take full control of the drone interrupts the UAV reception of the GCS control signals by breaking the communication link. The jamming attack results in the loss of communication between UAVs and the GCSs through the loss of control signals, which leads the drone to enter into a \textit{lost-link state} preventing the controller from operating correctly, and thereby causing unavailability of services.

\item \textbf{Denial of service (DoS):} An attacker may use the MAVLink vulnerability to flood the UAVs GCS communication channels with data; the network becomes interrupted, which leads to resource (UAV and GCS) unavailability. This form of attack is called a DoS attack. 
In such an attack, the control messages as well as the mission data is not properly received by the drone. As a result, the drone cannot remain in a stable state, and the mission is not executed appropriately.
If DoS attack is successful, it can result in a (M-I-T-M) attack. By conducting a MITM attack, the attacker sends unauthorized commands in an infinite loop to the UAV. This could effectively deny the communication between the GCS and the UAV, preventing the legitimate commands sent by the GCS from being treated by the UAV, as the drone would always be occupied by commands issued from the attacker.
A successful DoS attack against the drone makes it no longer responsive to the GCS, or vice versa, because of the violation of system availability.

Furthermore, in DDoS attack, an adversary sends a huge number of packets to the UAV or the GCS which causes a network congestion and prevents the UAV and the GCS from communicating with each other (failing to respond to commands).

\item \textbf{Flooding:} This attack works on the principle of flooding the network with a huge number of various packets to make it down. Generally, packets of types SYN, UDP, ICMP and Ping are used in this kind of attacks.
In \cite{kwon2018empirical}, a simulated attack, ICMP flooding attack, was performed to exploit the vulnerability of the MAVLink waypoint protocol. The intruder sends many ICMP request packets to both the GCS and the UAV during their mission. The GCS and the UAV are too overloaded, and thus, cannot respond to commands. As such, the UAV sensor values, the GCS mission commands were not appropriately transmitted.
Furthermore, the heartbeat message necessary for maintaining the connection between the UAV and the GCS is received after the target time because of the ICMP flooding attack. In this case, the UAV crashed without operating the fail-safe mode due to an error in the fail-safe mode.

\end{itemize}
\subsubsection{Authenticity attacks} Authenticity attacks try to make the GCS/UAV believe that falsified data is authentic. Authenticity of a MAVLink message can be hampered by fabricating malicious data to replace the legitimate data. Fabrication attacks include data fabrication and GCS spoofing.

\begin{itemize}
\item \textbf{Data fabrication:} To perform a fabrication attack, the adversary needs to have extended knowledge about how the GCS and the considered drone communicate, i.e., knows the protocol used by the drone and the GCS, which can be achieved by having performed eavesdropping and traffic analysis attacks. This attack violates the integrity of the control data and/or telemetry data, and can lead to a hostile takeover of the drone.
\item \textbf{GCS spoofing:} The UAV connects with the GCS via wireless links for data and control signal exchanges. However, since the wireless environment is open, an attacker could successfully spoof MAVLink commands. More specifically, a malicious attacker can send a false wireless control command to take over the UAV illegitimately. 
    
\end{itemize}

\begin{figure*}[tbh!]
\centering
\includegraphics[width=0.8\textwidth]{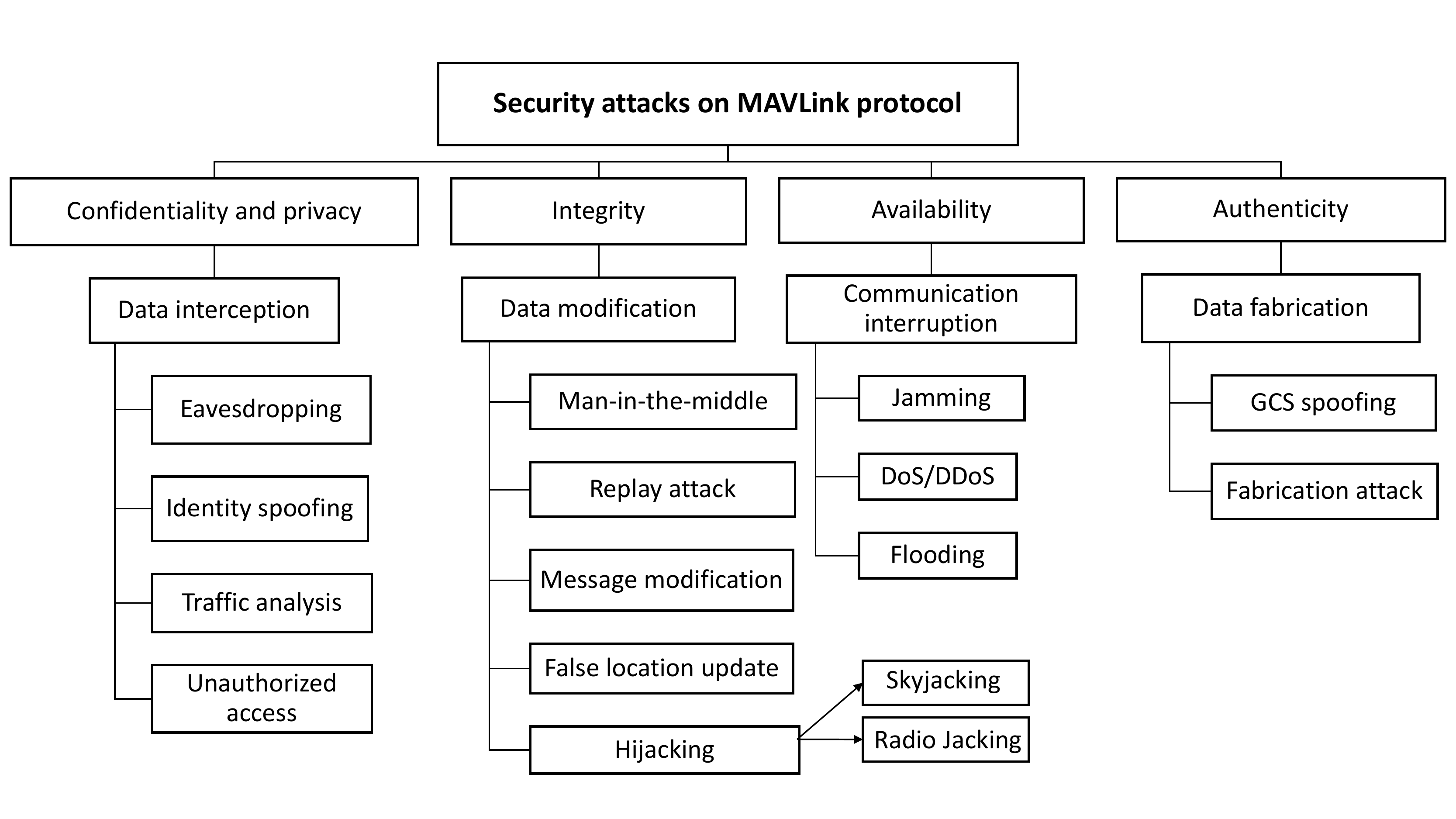}
\caption{Security threats and attacks against MAVLink Protocol}
\label{fig:sec-attacks}
\end{figure*}

Fig. \ref{fig:sec-attacks} summarizes the potential security threats against the MAVLink protocol, the identified attacks, and the corresponding violated security properties.

\section{MAVLink Security Solutions }
Despite the widespread use of the MAVLink protocol, it has security gaps and is prone to several attacks that result in critical threats and safety concerns. The protocol does not implement encryption and authentication mechanisms. Therefore, MAVLink is very prone to security threats and attacks. 

Several techniques and solutions have been already proposed to address the security issues described previously. In this section, we provide an overview and a classification of the proposed solutions for each security service as illustrated in Table. \ref{tab:sol}. Existing security solutions proposed for securing the MAVLink communication protocol can be classified into \textit{hardware} and \textit{software} approaches.

\begin{table*}[htb!]
\caption{Survey on the state-of-the-art on existing security solutions}
\label{tab:sol}
\resizebox{\textwidth}{!}{%
\begin{tabular}{|c|c|c|c|c|}
\hline
Category & \multicolumn{1}{c|}{Key method} & Proposed solution & Research focus & \multicolumn{1}{c|}{REF} \\ \hline
\multirow{2}{*}{Hardware based solution}& AES-CBC-MAC embedded in FPGA module & \multirow{2}{*}{Hardware} & \multicolumn{1}{c|}{\begin{tabular}[c]{@{}c@{}}Confidentiality and\\ authentication\end{tabular}} & \cite{shoufan2015secure} \\ \cline{2-2} \cline{4-5} 
 & Additional encrypted channel through Raspberry \textit{Pi}&  & \multicolumn{1}{c|}{\begin{tabular}[c]{@{}c@{}}Resume the control of UAV if\\ any attack was detected\end{tabular}}&  \cite{yoon2017security} \\ \hline
\multirow{19}{*}{Classical security approaches} & AES & \multirow{6}{*}{Symmetric key} & \multirow{9}{*}{Confidentiality} & \begin{tabular}[c]{@{}c@{}}\cite{ pike2013keynote}, \cite{smavlink}\end{tabular} \\ \cline{2-2} \cline{5-5} 
 & Galois Embedded Crypto Library &  &  & \cite{podhradsky2017improving} \\ \cline{2-2} \cline{5-5} 
 & AES-ECB, AES-CBC &  &  & \cite{han2017authentication} \\ \cline{2-2} \cline{5-5} 
 & \begin{tabular}[c]{@{}c@{}}Rabbit stream cipher, XXTEA\\ stream cipher, and Salsa20 stream cipher\end{tabular} &  &  & \cite{marty2013vulnerability} \\ \cline{2-2} \cline{5-5} 
 & \textit{RC5} &  &  & \cite{butcher2013securing} \\ \cline{2-2} \cline{5-5} 
 & Caesar cipher &  &  & \cite{rajatha2015authentication,Hamsavahini} \\ \cline{2-3} \cline{5-5} 
 & RSA and ECC & Asymmetric key &  & \cite{han2017authentication} \\ \cline{2-3} \cline{5-5} 
 & \begin{tabular}[c]{@{}c@{}}probabilistic selective data\\ encryption\end{tabular} & \multirow{2}{*}{\begin{tabular}[c]{@{}c@{}}Identity-based encryption\\ (IBE)\end{tabular}} &  & \cite{haque2018new} \\ \cline{2-2} \cline{5-5} 
 & (IBE-Lite) &  &  & \cite{lin2018security} \\ \cline{2-5} 
 & Private key & Digital signature & \multirow{4}{*}{Integrity} & \cite{bian2013secure} \\ \cline{2-3} \cline{5-5} 
 & \begin{tabular}[c]{@{}c@{}}Authenticated encryption\\ algorithm\end{tabular} & \multirow{3}{*}{Symmetric key} &  & \cite{altawy2017security} \\ \cline{2-2} \cline{5-5} 
 & \begin{tabular}[c]{@{}c@{}}Message Authentication Code( Poly1305)\\ and authenticated encryption algorithm\\ Galois/Counter Mode (GCM)\end{tabular} &  &  & \cite{marty2013vulnerability} \\ \cline{2-2} \cline{5-5} 
 & \begin{tabular}[c]{@{}c@{}}Message Authentication Code\\ (MAC)\end{tabular} &  &  & \cite{zouhri2017new} \\ \cline{2-5} 
 & \begin{tabular}[c]{@{}c@{}}Strong authentication based\\ solution\end{tabular} & Symmetric key & Availability & \cite{shakhatreh2018unmanned,shoufan2015secure} \\ \cline{2-5} 
 & AES-GCM & \multirow{3}{*}{Symmetric key} & \multirow{5}{*}{Authenticity} & \cite{verup2016security,marty2013vulnerability} \\ \cline{2-2} \cline{5-5} 
 & Caesar Cipher &  &  & \cite{rajatha2015authentication,Hamsavahini} \\ \cline{2-2} \cline{5-5} 
 & \begin{tabular}[c]{@{}c@{}}Message Authentication Code\\ (MAC)\end{tabular} &  &  & \cite{shoufan2015secure} \\ \cline{2-3} \cline{5-5} 
 & \begin{tabular}[c]{@{}c@{}}Elliptic Curve Cryptography\\ (ECC)\end{tabular} & Asymmetric key &  & \cite{han2017authentication} \\ \cline{2-3} \cline{5-5} 
 & \begin{tabular}[c]{@{}c@{}}The signature represents the first 48\\ bits of an SHA-256 hash of the\\ secret key, header, payload,\\ CRC, link ID, and timestamp\end{tabular} & Digital signature &  & \cite{mavlinkv2} \\ \hline
\multirow{8}{*}{\begin{tabular}[c]{@{}c@{}}Intrusion Detection System\\   (IDS)\end{tabular}} & Behavior rule-based solution & \multirow{2}{*}{\begin{tabular}[c]{@{}c@{}}Rule-based specification\\ detection\end{tabular}} & \begin{tabular}[c]{@{}c@{}}Detect and guard a UAV system\\ against cyber-attacks\end{tabular} & \cite{mitchell2014adaptive} \\ \cline{2-2} \cline{4-5} 
 & Behavior rule-based solution &  & \begin{tabular}[c]{@{}c@{}}Evaluate the behavior of\\ attacks that target UAV\end{tabular} & \cite{kim2012cyber} \\ \cline{2-5} 
 & \begin{tabular}[c]{@{}c@{}}UAV behavior based fight\\ commands\end{tabular} & Signature-based detection & Authentication & \cite{shoufan2017continuous} \\ \cline{2-5} 
 & \begin{tabular}[c]{@{}c@{}}Statistical method (recursive least squares\\ technique)\end{tabular} & \multirow{3}{*}{Anomaly-based detection} & \begin{tabular}[c]{@{}c@{}}UAV real-time monitoring\\ system\end{tabular} & \cite{birnbaum2016unmanned} \\ \cline{2-2} \cline{4-5} 
 & Belief-based threat estimation &  & \begin{tabular}[c]{@{}c@{}}Protect UAVs from attacks\\ targeting data integrity\end{tabular} & \cite{sedjelmaci2016detect} \\ \cline{2-2} \cline{4-5} 
  & Neural network and fuzzy learning algorithm &  & \begin{tabular}[c]{@{}c@{}}Protect UAVs against a distributed\\ denial-of-service (DDoS) attacks\end{tabular} & \cite{rani2016security} \\ \cline{2-2} \cline{4-5}
  & Support Vector Machine (SVM) algorithm &  & \begin{tabular}[c]{@{}c@{}}Detect cyber-attacks that\\ target autonomous avionic systems\end{tabular} & \cite{casals2013generic} \\ \cline{2-2} \cline{4-5}

 & Bayesian game model &  & \begin{tabular}[c]{@{}c@{}}Protect UAV-aided network\\ against lethal attackers\end{tabular} & \cite{sedjelmaci2017intrusion} \\ \cline{2-5} 
 & \begin{tabular}[c]{@{}c@{}}Rule-based detection and\\ SVM-based anomaly detection\end{tabular} & \multirow{2}{*}{Hybrid-based detection} & Identify cyber-attacks & \cite{sedjelmaci2018hierarchical} \\ \cline{2-2} \cline{4-5} 
 & \begin{tabular}[c]{@{}c@{}}Signature-based anomaly detectors and\\ residual-based anomaly detectors.\\ Bayesian network to estimate possible attacks\end{tabular} &  & Detects GPS spoofing attacks & \cite{muniraj2017framework} \\ \hline
\multirow{4}{*}{\begin{tabular}[c]{@{}c@{}}New emerging security\\ solutions\end{tabular}} & \multirow{4}{*}{Blockchain} & \multirow{4}{*}{Blockchain} & \begin{tabular}[c]{@{}c@{}} Data integrity, \\trusted source ,\\ accountability, and\\ resilient backend\end{tabular} & \cite{liang2017towards} \\ \cline{4-5} 
 &  &  & \begin{tabular}[c]{@{}c@{}}Secure the communication\\ among UAVs\end{tabular} & \cite{kapitonov2017blockchain} \\ \cline{4-5} 
 &  &  & \begin{tabular}[c]{@{}c@{}}Securely relay drone\\ information\end{tabular} & \cite{sharma2017socializing} \\ \cline{4-5} 
 &  &  & Security and privacy & \cite{dissiminaationIoD} \\ \hline
\end{tabular}%
}
\end{table*}

\subsection{Hardware-based solution}
Several embedded and hardware security solutions have been introduced to secure the MAVLink protocol. In \cite{shoufan2015secure}, a lightweight hardware-based solution is proposed to secure the communication between the GCS and the drone. An FPGA module connected to the drone embeds the symmetric key cryptography function: AES-CBC-MAC was used to encrypt and authenticate both commands and payload data communicated between the drone and the GCS. However, the hardware solution negatively affects system performance and power consumption due to the extra hardware weight. 
In \cite{yoon2017security}, the authors proposed the idea of an additional encrypted communication channel to improve UAV data security through Raspberry \textit{Pi}. This channel was designed to resume the control of the UAV if any attack was detected on the drone. However, this hardware solution induces delays between the GCS and the Raspberry \textit{Pi} and increases the CPU usage on the Raspberry \textit{Pi}. 

\subsection{Software based solution}
 We may here further distinguish among several approaches: the classical security approaches, the Intrusion Detection System (IDS) based approaches, and the new emerging Blockchain technology based approaches.
\subsubsection{Classical security approaches} 
    This category of solutions groups cryptographic-based approaches used in the context of MAVLink to address the main security services. We review the proposed cryptographic solutions and discuss their main advantages and shortages relatively to each security service.
    
\paragraph{Confidentiality solutions} To ensure confidentiality of both control signals and telemetry data, MAVLink data has to be encrypted before being sent. However, it is important to use both a strong key and a powerful cryptographic algorithm to mitigate the vulnerabilities of the MAVLink protocol in terms of confidentiality. Cryptographic algorithms are classified into two main classes: symmetric and asymmetric cryptographic solutions.

\textbf{(\textit{i.})\paragraph{ Symmetric key solutions:}}
Each entity in the system shares the same cryptographic keys with all other entities. The main benefits of symmetric key schemes are their easy implementation, fast design, and low computation requirements. In fact, they use the same key to encrypt and decrypt data, which makes it appropriate for limited-resource drones.

In \cite{ pike2013keynote}, the authors introduced SMACCMPilot, a secure UAV project based on the MAVLink protocol. SMACCMPilot refers to GIDL as the application level protocol. GIDL uses AES to encrypt the MAVLink payload, header, and CRC. The community of the MAVLink protocol developers are currently discussing a secure version of MAVLink (sMAVLink)\cite{ smavlink}. sMAVLink has the same encryption algorithm as GIDL but encrypts only the payload, which makes the MAVLink packet structure untouched. To the best of our knowledge, sMAVLink has not yet been implemented.

In \cite{podhradsky2017improving}, the authors proposed an encrypted radio control link based on Galois Embedded Crypto library with the openLRSng open-source radio project for securing communication links among open source UAV systems. The proposed solution uses the symmetric key produced by a trusted third-party entity and manually hard-coded in the autopilot. This approach is not efficient and may lead to security vulnerabilities and restricts its feasibility.
In \cite{han2017authentication}, the authors chose AES-ECB and AES-CBC to protect command messages.

In \cite{zouhri2017new}, the authors proposed a transfer protocol, which provides confidentiality service and data transfer service between the drone and GCS. User data are encrypted with a key, derived during the initialization phase to guarantee data confidentiality, 

The work in \cite{marty2013vulnerability} presented four suitable cryptographic implementations that may be able to mitigate the confidentiality vulnerabilities presented in the MAVLink protocol using
strong symmetric-key encryption algorithms. The proposed algorithms are Rabbit stream cipher, XXTEA stream cipher, and Salsa20 stream cipher. They both encrypt MAVLink messages rapidly while preserving the confidentiality of communication among the GCS and UAV. 

Subsequently, an encryption mechanism \textit{RC5} is used in \cite{butcher2013securing} to secure the MAVLink communication protocol.
In \cite{rajatha2015authentication},\cite{Hamsavahini}, the Caesar cipher cryptography algorithm is used for data encryption of MAVLink messages between the ground station and the Micro Aerial Vehicles (MAV). However, in this method, the secret key is sent as a plain text to the drone, during the establishment phase. The key could easily be detected which
breaks the whole security system. Moreover, the Cesar encryption
algorithm used in this method is proved to be insecure and is vulnerable to cryptanalysis.

\textbf{(\textit{ii.})\paragraph{ Asymmetric key solutions:}}
Another form of encryption is public-private key cryptography, also known as asymmetric encryption. It uses a couple of public/private keys. The public key is used for data or information encryption, and the private key, only known to the receiver, is used for the decryption process. The advantages of these approaches are their flexibility, scalability and skey management efficiency. However, these solutions can cause severe computational, memory, and energy overhead which are not suitable for constrained devices. RSA and ECC algorithms were used in  \cite{han2017authentication} to encrypt of Aerial Robotics Communication.

\textbf{(\textit{iii.})\paragraph{ Identity-based encryption (IBE):}}
The problem of public key cryptography resides in its dependency on the third-party authority that issues the certificates. To overcome the scalability and complexity issues, IBE is proposed by suggesting the idea to use known information that uniquely identifies users (e.g., phone number, email, etc.) as their public keys for data encryption and thus eliminates the necessity for certificates and Public Key Infrastructures (PKI).
Despite the clear achievement in scalability and efficiency, IBE needed further refinements to become lightweight and consequently viable for the use in resources constrained devices such as UAVs.

The contribution in \cite{haque2018new} is twofold. First, a hierarchical architecture has been designed for the UAV networks using identity-based encryption and bilinear pairing over elliptic curve cryptography (ECC) without compromising system security.
Second, a lightweight cryptographic primitive is proposed using a probabilistic selective data encryption technique. The proposed method improves system performance and increases the efficiency of the transmitted message without affecting security. Stenography or data watermarking technique is used to reduce overheads and increase message confidentiality. In \cite{lin2018security}, a lightweight IBE scheme (IBE-Lite) is specially designed for resource-constrained IoD architecture. The proposed scheme facilitates the secure sharing of drones' data.

\paragraph{Integrity solutions}    
Integrity can be ensured using signature, hash functions, message authentication code (MAC) and authenticated encryption cryptographic primitives \cite{rodday2016exploring, benkraoudacyber}.

In \cite{bian2013secure}, the authors proposed to add a digital signature to the data packet using the UAV private key.

According to \cite{altawy2017security}, the authors proposed to use authenticated encryption cryptographic mechanisms to enforce the integrity of the data.

In \cite{marty2013vulnerability}, the authors addressed two cryptographic implementations that can mitigate integrity vulnerabilities presented in the MAVLink protocol: the Poly1305 Message Authentication Code (MAC) and the Galois/Counter Mode (GCM). The GCM implementation adds an authentication code along with the ciphertext and the initialization vector. The authentication code is used to create an authentication tag that is used as a method for validating the message integrity. Poly1305 is a message authentication code (MAC) used to verify data integrity.

However, using Poly1305 or GCM increases the packet size due to the added padding, which increases the latency in UAV communications and reduces the energy efficiency of the autopilot processor. Moreover, Galois/Counter Mode is the most computation-extensive mechanism because it authenticates and encrypts every message using a slower method.

In \cite{zouhri2017new}, the authors proposed a data transfer and confidentiality services between the ground station and the drone. This protocol ensures the message integrity using a message authentication code (MAC) function using an alternative key computed based on the master key.

\paragraph{Availability solutions} Protecting the UAV against malicious availability attacks is important to succeed in UAV missions. 

The contribution in \cite{sedjelmaci2016detect} consists of proposing an estimation model based on estimated beliefs to detect the existence of a system threat. This work includes specific detection policies to maintain the availability of UAV network.

Authors in \cite{shakhatreh2018unmanned,shoufan2015secure} surveyed the potential DoS attacks that could cause serious availability issues in UAV systems. They also proposed strong authentication based solutions to mitigate these attacks. 

\paragraph{Authenticity solutions} Authentication is essential to make sure that the GCS is controlling an authorized drone not a fake one, and that the UAV is sending its state or accepting commands from a legitimate GCS not from a hacked/fake ground station. In other words, authentication enables the UAV and the GCS to guarantee that they communicate with each other. Authentication techniques used in UAV networks are Symmetric key solutions, Asymmetric key solutions, and Digital signature.

\textbf{(\textit{i.})\paragraph{ Symmetric key solutions:}}
A symmetric encryption algorithm AES-GCM is used in \cite{verup2016security,marty2013vulnerability} to ensure the authenticity of the transmitted signals. In \cite{rajatha2015authentication,Hamsavahini}, the authors tackled the problem of authentication using the Caesar Cipher method in the MAVLink protocol. To guarantee a secured communication between the UAV and the GCS, the drone is authenticated at the beginning of the communication using Cesar Ciphering. Thus, GCS data will not be received by the UAV unless it gets authenticated, through sending an encrypted text similar to the one produced by the GCS. 

The Message Authentication Code (MAC) algorithm is used to provide message authentication by using a symmetric key encryption technique. Authors in \cite{shoufan2015secure} claimed that attaching the MAC to each MAVLink payload could be suitable to verify the authenticity of the message.

\textbf{(\textit{ii.})\paragraph{ Asymmetric key solutions:}}
The authors in \cite{han2017authentication} employed the use of Elliptic Curve Cryptography (ECC) for ensuring authentication between the GCS and the UAV. All feedback messages are encrypted with the drone's private key, which means that only the drone's public key can decrypt these messages. This procedure ensures that the receiver authenticates the origin of the received messages. On the other hand, the GCS's public/private keys guarantee that the UAV recognizes the message sent by the GCS so that the drone is authenticated.

\textbf{(\textit{iii.})\paragraph{ Digital signature:}} As we stated before, a MAVLink version 2.0 has been designed to support the packet signing mechanism and to bring security to MAVLink communications \cite{mavlinkv2}.
The signature represents the first forty eight bits of an SHA-256 hash of the
secret key, payload, link ID, header, CRC, and timestamp. The 13 bytes (including signature, link ID and timestamp) must be appended at the tail of a MAVLink 2.0 message to ensure that messages are sent by trusted sources.

Despite packet signing has backward-compatibility and best portability as compared to other security alternatives, developers still need to upgrade their autopilots to MAVLink v2.0 to support packet signing \cite{davanian2017diversity}.

However, adding packet signing security solution is not free. The cost consists in several factors including the computational overhead of 26 microseconds per packet, the increase of firmware size (packet signing code is 812 bytes) and the increase in power consumption due to the higher CPU and network traffic usages. This leads also to increase the communication time \cite{davanian2017diversity}.

\subsubsection{Intrusion Detection System (IDS)} It is essential to protect the UAV system against attackers by allowing to detect possible cyber attacks against the drone communication system.
Intrusion Detection Systems (IDSs) are typically deployed to monitor the incoming communication, supervise and identify indications of abnormal activity or behavior. Some works in \cite{mitchell2014adaptive,kim2012cyber,shoufan2017continuous,birnbaum2016unmanned,sedjelmaci2016detect,sedjelmaci2017intrusion,sedjelmaci2018hierarchical,muniraj2017framework,rani2016security,casals2013generic} have discussed the area of intrusion detection systems (IDSs) for UAVs. In what follows, we survey these existing studies based on the categories of the techniques proposed for intrusion detection including, the rule-based specification detection, anomaly-based detection,  hybrid-based detection, and signature-based detection,.

    \paragraph{Rule-based specification detection} This method is based on comparing the behavior of UAVs against a set of specified rules based on the expected behaviors of UAVs. 

In \cite{mitchell2014adaptive}, Mitchell and Chen proposed the BRUIDS intrusion detection mechanism, which aims to detect and protect a UAV system against  security threats. BRUIDS is a rule-based specification detection technique for intrusion detection of compromised UAVs. The authors proposed a set of behavioral rules related to cyber-attacks constructed based on defined attack models to build a model of a normal UAV behavior.

Kim et al. \cite{kim2012cyber} evaluated the behavior of attacks that target UAV. They proposed a behavior rule-based intrusion detection system for UAVs, in which the rules are specified according to these malicious anomalies to model a normal UAV behavior. 

     \paragraph{Signature-based detection} This method refers to the detection of known attacks based on predefined known signatures, features, and patterns. Such signatures are used to compare patterns with captured events to recognize and ensure the detection of possible intrusions. Although signature-based detection can easily detect known attacks, it is difficult to detect unknown or new attacks for which no pattern is available.

Authors in \cite{shoufan2017continuous} proposed a technique (behaviometrics) for continuous authentication of data command transmitted by a ground station to the drone based on UAV behavior. The drone's behavior is specified using a set of flight commands, which are considered later as a unique signature to identify authorized UAVs and detect malicious commands stemmed from attackers.

  \paragraph{Anomaly-based detection} Anomaly-based mechanism defines normal and baseline features to build a model of normal behavior profile and to follow any possible variation from the normal behavior. Anomaly detection usually uses statistic analysis, machine learning techniques, and game theory to enhance the detection of anomaly behavior of a monitored node and unknown attacks. The key benefit of this mechanism lies in its ability to detect new or unknown attacks when there are no predefined signatures of the unknown attacks.

Authors in \cite{birnbaum2016unmanned} presented another IDS, where they proposed a prototype of UAV real-time monitoring system to control avionics and flight controller systems. Their approach adapts the recursive least squares technique to estimate UAV navigation sensor, controller parameters, and other related parameters. Through the application of this statistical method, the IDS can identify the system's parameters values, and the anomaly is detected whenever the monitored parameters deviate from their expected values during the flight.

In \cite{rani2016security}, the authors suggested an intrusion detection system based on Neural network and fuzzy learning algorithms to protect UAVs against a distributed denial-of-service (DDoS) attacks.

Authors in \cite{casals2013generic} developed an anomaly detection scheme based on Support Vector Machine (SVM) algorithm to detect cyber-attacks that target autonomous avionic systems. 

The authors in \cite{sedjelmaci2016detect} proposed and implemented an intrusion detection system to protect UAVs from attacks targeting data integrity. The detection of these attacks relies on a belief-based threat estimation model to reduce false positive and false negative rates. 

Besides, the authors in \cite{sedjelmaci2017intrusion} proposed a collaborative intrusion detection framework, named Security Game Framework (SGF), to protect UAV-aided network against lethal attackers. SGF is formulated based on the Bayesian game model to detect attacks accurately.

   \paragraph{Hybrid-based detection} It takes the advantages of both rule-based, anomaly detection, and signature-based  and combines them to catch known and/or unknown attack signatures and abnormal events.
    
   In \cite{sedjelmaci2018hierarchical}, authors investigated the use of an intrusion detection. The IDS uses an SVM learning algorithm to classify threats while monitoring the behavior of UAVs. A sequence of detection policies related to each cyber-attack is proposed based on a hybrid approach (rule-based detection and SVM-based anomaly detection) to model a normal UAV behavior. 

The authors in \cite{muniraj2017framework} proposed a framework based on IDS that detects GPS spoofing attacks onboard the UAV. The IDS uses the attack-signature-based anomaly detectors as well as residual-based anomaly detectors. The Bayesian network takes anomaly detectors outputs as evidence to estimate a possible attack through Bayesian inference.

\subsubsection{New emerging security solutions}
Blockchain is an emergent technology that can be efficiently used  solve the aforementioned security issues of MAVLink.
Blockchain is originally used for recording financial transactions between entities in a distributed and decentralized manner. The transaction is verified collaboratively using trusted entities in the network, thus eliminating the need for a controlling authority. Moreover, transactions are stored on the Blockchain, which makes tampering with data extremely challenging because Blockchain relies on fully distributed cryptographic techniques. In this way, any modification on these transactions can be easily detected.

All these advantages led several researchers to consider this technology to deal with security issues in UAV network since Blockchain provides privacy, integrity, accountability, authorization, authentication, confidentiality, identity hiding and non-repudiation.

In \cite{liang2017towards}, the authors included Blockchain and cloud storage in their framework to guarantee the UAV data integrity. This idea addresses the following objectives: trusted source, timing, and data integrity, accountability and resilient backend.

In \cite{kapitonov2017blockchain}, Blockchain is used to secure the communication among UAVs as they collaborate to make cooperative decisions and exchange data.

Sharma et al. \cite{sharma2017socializing} exploited the Blockchain features to securely relay drone information, especially in ultra-dense environments.

Furthermore, authors in \cite{dissiminaationIoD} presented a system model based on the public Blockchain technology which provides security and privacy to the IoD network. 

The approach proposed by \cite{GARCIAMAGARINO201972} relies on Blockchain principles to identify compromised UAVs using trust rules and detect wrong information when a UAV is hijacked.

\section{Literature Review} \label{sec:lit}
The MAVLink protocol has attracted the research community, and several contributions were proposed in the literature. Some of these works proposed some extensions and enhancements to the protocol (e.g., \cite{Dietrich2016}, \cite{ZACARIAS2016}, and \cite{ERDELJ2017}), while some other works presented the integration of the MAVLink protocol with the cloud and the Internet of Things (e.g., \cite{Aljehani2017}, \cite{KOUBAA201946}). Furthermore, recent works also addressed how to use MAVLink for autonomous agents and swarms (e.g. \cite{Daniel2018}, \cite{Braga2018}, \cite{Daniel2018}). In this section, we present an overview of the recent research contributions that dealt with the MAVLink protocol. 

\subsection{Extensions and enhancements}
Recently, several research studies proposed extensions to the MAVLink protocol. In particular, the authors in \cite{Dietrich2016}, \cite{ZACARIAS2016}, and \cite{ERDELJ2017} extended the MAVLink protocol to support the multi-drones' cooperation. 

In \cite{Dietrich2016}, the authors defined a new set of messages and data structures to manage a swarm and to enable drone-to-drone communication. The proposed messages are divided into two groups: swarm formation set and swarm maintenance set. In total, the authors developed six enumerations, six new commands, and thirty-three new messages. However, the authors did not validate their approach neither with a simulation nor experimental implementation. The concept of groups was introduced into the MAVLink protocol by adding two identifiers: 1) Group ID (group-wide), and 2) Group Member ID (group-internal). For the swarm maintenance, two factors were considered: drone replacement and recharging. The drone replacement is the process of finding a suitable alternative drone, then move all the mission-related data to that new drone, and finally the physical and logical replacement of the old drone with the new drone to finish the mission. The drone replacement is needed in several cases such as when the flight duration is undefined or longer than the capacity of the battery, or when the drone encounters some hardware problems. The recharge is a specific operation to extend the mission duration because of the limited battery capacity.

In \cite{ERDELJ2017}, Erdelj et al. also proposed additional new MAVLink messages and commands support collaboration between drones in multi-UAV scenarios. The paper proposed an approach to ensure the continuity of the drone service, which means that when a drone has to leave the mission, it is \textit{immediately} replaced with another. A proof-of-concept simulation was presented to show the effectiveness of the proposed algorithm. Also, the authors analyzed the performance of their system in terms of the total number of bytes induced by the collaboration.



\begin{figure}[htbp]
	\centering
	\includegraphics[width=0.49\textwidth]{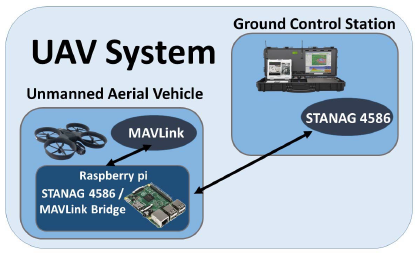}
	\caption{Bridge between MAVLink and STANAG 4586 \cite{Rodrigues17}}
	\label{mavlink-stanag}
\end{figure}

The work in \cite{Rodrigues17}, presented a system that translates MAVLink messages to STANAG 4586 standard \cite{marques2012stanag}. STANAG 4586 is a complete standard aiming at having NATO interoperability between UAVs from different countries. A bridge between MAVLink and STANAG 4586 created in a Raspberry Pi computer to make it easy to integrate with any UAV (Fig. \ref{mavlink-stanag}). This work aimed to allow any GCS compatible with STANAG 4586 to be able to operate with any MAVLink-based UAV to accomplish the interoperability between the UAVs of the NATO's member. The proposed system was tested using the SITL simulator. Only important messages were translated such as waypoints messages because STANAG 4586 has a large set of messages. The messages were received successfully by the UAV, and the change on the parameters (long, lat, and alt) was not significant.

\subsection{Cloud and IoT Integration}

The use of Internet-of-Things to communicate, manage and control multiple drones has gained a lot of interest.

In \cite{KOUBAA201946}, the authors proposed Dronemap Planner, which is a cloud-based system to manage drones over the Internet and to offload heavy computation from the UAV to the cloud (e.g., Image Processing). The authors used the MAVLink protocol to send information from the drone to the cloud. The cloud forwards UAV data to corresponding users, which also send their commands to the drones through the cloud. The paper provided a complete implementation of the cloud-based management system and demonstrated how to monitor and control drones over the Internet effectively. 

The work in \cite{Aljehani2017} proposed a multi-UAV system for tracking and scanning missions in disaster response applications. A UAV is turned into an IoT device by embedding an 4G dongle with the drone autopilot. The MAVLink protocol is used for the communication between the UAV and the GCS. The authors proposed to use APM software on Windows 2016 OS running on a cloud server (Elastic Compute Cloud (EC2)). Experiments on a scanning mission for an area of size 0.16 km with real drones were conducted on a university campus. The missions were executed on EC2 and MAVLink through UDP and TCP protocols to exchange data and receive commands from the ground station.  In addition, this paper analyzed the behavior of the MAVLink protocol using both UDP and TCP connections through 4G network.

In \cite{Borey2018}, the authors proposed an IoT architecture of different types of systems including UAVs, sensor devices and mobile phones. The communication between UAVs and the server was performed using the MAVLink protocol. 

The work in \cite{Hadiwardoyo2018} aimed to create an framework where the drones might be used to support the communication in areas having no or limited infrastructure. The focus is on the point-to-point connection between drone and car in an ad-hoc communication. The position information is received from the UAV's GPS using MAVLink protocol, which allows the communication between the drone's hardware and the Raspberry Pi.



\subsection{Simulation and Modeling}
Several research works proposed simulation frameworks and evaluated the performance of MAVLink-based unmanned aerial systems. 

The work in \cite{Atoev2017} analyzed the data loss and network latency  in the communication channel between the ground station and the autopilot. The work in \cite{smigielski17} describes a simulator for MAVLink-based UAVs. The proposed visual simulator combines SITL simulator which uses MAVLink for the communication with a scene model generated in OpenGL environment which allows visualizing the actions executed by the simulated UAV. 

The work in \cite{SUN2018} designed a Hardware-In-the-Loop testbed to test and simulate a control system for tail-sitter UAV. MAVLink is used to exchange messages between the vehicle and the GCS. The work in \cite{FABRA2018}, introduced a new simulator called ArduSim which allows to control UAV flights in real-time and manage the communication between multiple UAVs. The MAVLink is used to send the control messages via TCP. In \cite{Pannocchi2018}, the authors proposed a modular hardware-in-the-loop (HITL) simulation framework for multi-vehicle autonomous systems. The proposed simulation framework is compatible with any ground control station that operates MAVLink protocol over UDP.


The work in \cite{NIELSEN2018} addressed the development of a methodology for automated modeling of collaborative underwater vehicles. The MAVLink protocol is used to communicate with the Pixhawk autopilot and the Odroid-XU4 single board computer of the robot.

The work in \cite{Mason2018} proposed a framework that combines different verification methods (simulation, symbolic, and statistical and model checking) to analyze the different stages of drone development. The proposed framework is composed of 3 main components: (1) Simulator: to specify the physical model of the drone, (2) formal executable language: used to specify the drone behavior and environment, (3) statistical model checking algorithm: used to analyze the system behavior. The communication uses the MAVProxy ground station that acts as a proxy that forwards the MAVLink commands to the drone. 


\subsection{Applications}

The MAVLink protocol is used as a communication protocol in a wide range of UAV applications, ranging from agriculture, construction, and environment monitoring. In what follows, we present an overview of some applications. 

In \textbf{Agriculture applications}, the work in \cite{Mogili2018} proposed a system for crop monitoring through a multispectral camera mounted on a UAV. The MAVLink protocol is used for the wireless communication between the UAV to the ground station to send the multispectral crop images through telemetry. In the context of \textbf{environmental monitoring applications}, the work in \cite{Schill2018} proposed an autonomous underwater vehicle system for cooperative environmental sensing. The MAVlink protocol is used to handle the communication between the modules, in addition the base station workstation. 
In \cite{Kabaldin2018}, the authors developed an autonomous drone for the monitoring of oil and gas pipelines. All the data are transmitted using the MAVLink protocol, which ensures reliable communication within a 5-km radius. For \textbf{construction inspections}, the work in \cite{Freimuth2018}, proposed to integrate UAVs for inspection purposes and used MAVLink for communication between Ground Control Stations and UAVs. For \textbf{Maritime application}, the work in \cite{Leopoldo2018} proposed an architecture for maritime surveillance using battery-powered drones. The MAVLink protocol was used to exchange data between the autopilot and the flight duration enhancement system. 


In addition, several recent research works addressed disaster management applications, such as \cite{Riviere2017} and \cite{Menegol2018}. In \cite{Riviere2017}, the authors developed a fleet of UAVs for research and rescue applications. Rescue operations are sent from mobile phone to the drones through the MAVProxy ground station. 
The work in \cite{Menegol2018} proposed an agent-hardware integration architecture for search and rescue operations. The proposed architecture embeds JaCaMo agents on UAVs. The UAV's flight controller exchanges data via MAVLink protocol.


\subsection{Autonomous Agents}

There are several efforts have been made to make the UAVs autonomous. In \cite{Daniel2018}, authors proposed an architecture to provide the UAV with the capability of locating itself using computer vision, modeling its environment, and planning and executing a 3D trajectories.  The work was successfully tested with Solo from 3D Robotics which compatible with MAVLink. Gstreamer is used to receive the video feed, and Dronekit, which is compatible with MAVLink, is used to gain access to the vehicle. In \cite{ALSHARMAN2018}, authors deal with precision landing problem for UAVs.
As the GPS quality reduced when the drone gets close to the ground, this paper proposed using low cost adaptive fuzzy multi-sensor data fusion architecture. PX4FLOW sensor is used to get an accurate velocity measurement and recognize the moving features. PX4FLOW data was acquired using the MAVlink protocol. The work in \cite{Acevedo2018} presents a system designed for multicopters to enable them from autonomous landing on a moving object. The presented system is based on visual tracking and landing of a marker on the object. The MAVLink protocol is used for the communication between the autopilot simulator and the rest of the system. The work in \cite{PRIMATESTA2018} proposed an algorithm to quantify the risk in a population and to find the optimal path with minimum risk.  The proposed algorithm consists of two phases: first, identify the risk in a specific area by generating a risk map based on factors like population density, no-fly zones, sheltering, and obstacles. The second phase is the path planning algorithm to search for the best path that minimizes the risk. The communication between the autopilot and ROS is done using MAVLink protocol.


The work in \cite{Mathias2016} proposed an algorithm for offline path planning in a static environment. The proposed algorithm was tested with real UAV. The algorithm runs on a single-board computer onboard (Odroid XU4) To allow a completely autonomous flight. The proposed system was tested using Iris+, a ready-to-fly quadcopter from 3DRobotics to demonstrate the efficiency of the proposed system in practice.  The Pixhawk flight control board is used to control the Iris+. The communication with the Pixhawk is done using MAVLink.

\subsection{Swarm Control}

In \cite{Braga2018}, authors presented an algorithm to avoid collisions for a swarm of UAVs. The authors used small quadrotors with diameter 250mm equipped with GPS and distance sensors. An embedded Linux computer (Raspberry \textit{Pi}) and Pixhawk autopilot board are used to control the UAVs. The control algorithm runs on the Raspberry \textit{Pi} and is implemented as C++ package in the ROS platform.  The MAVLink protocol is used for the communication between the Raspberry Pi and the Pixhawk.

The work in \cite{YUAN2017} proposed a decentralized control algorithm to be executed by each drone on the local onboard computer (Raspberry \textit{Pi}). The proposed algorithm implemented as Python scripts, and in each iteration after computing the desired velocity, they sent to the flight control through MAVLink using DroneKit Python interface.


\section{MAVLink Related Software} \label{sec:ardu}

There several types of software that support the MAVLink protocol, including (\textit{i.}) ground stations, (\textit{ii.}) Simulation frameworks. In the following subsections, we provide the main candidates for each category.   




\subsection{Supported Ground Stations}

A Ground Station is a software that communicates with the micro-vehicle through a serial or network interface by exchanging MAVLink messages. The communication can either take place over a serial port generally through a telemetry device or through a network interface inside a wireless local area network using UDP or the protocol. The advantage of telemetry devices is that they allow more extended communication range than traditional WLAN technologies, and can reach up to 5 km in range. This section presents the most common available ground stations. Table \ref{tab:GCS} shows a comparison of each GCS software discussed in this section.

\begin{itemize}

    \item \textbf{QGroundControl:} The most commonly used ground station is QGroundControl \cite{Qground}. This ground station was natively developed in C++ and also has a wrapper package for Android. It fully supports the MAVLink protocol in addition to Ardupilot and PX4 powered vehicles. QGroundControl has several functionalities, including defining and planning autonomous missions, full control of the vehicle, graphical visualization of the map and location tracking of the vehicle through its GPS coordinates. It also provides support for video streaming and changing the internal parameters of the autopilot, in addition to the calibration of the sensors of the autopilot. QGroundControl runs on different platforms, namely, Windows, Mac OS, IOS, and Android devices.
    
    \begin{table*}[htbp]
\centering
\caption{An overview comparison between GCS software}
\resizebox{\textwidth}{!}{%
\begin{tabular}{|c|l|l|l|l|l|l|l|}
\hline
\rowcolor[HTML]{EFEFEF} 
{\color[HTML]{333333} \textbf{GCS software}} & \multicolumn{1}{c|}{\cellcolor[HTML]{EFEFEF}{\color[HTML]{333333} \textbf{\begin{tabular}[c]{@{}c@{}}Free/\\ commercial\end{tabular}}}} & \multicolumn{1}{c|}{\cellcolor[HTML]{EFEFEF}{\color[HTML]{333333} \textbf{Interface}}} & \multicolumn{1}{c|}{\cellcolor[HTML]{EFEFEF}{\color[HTML]{333333} \textbf{\begin{tabular}[c]{@{}c@{}}Supported\\ Autopilots\end{tabular}}}} & \multicolumn{1}{c|}{\cellcolor[HTML]{EFEFEF}{\color[HTML]{333333} \textbf{Platforms}}} & \multicolumn{1}{c|}{\cellcolor[HTML]{EFEFEF}{\color[HTML]{333333} \textbf{\begin{tabular}[c]{@{}c@{}}MAVLink\\  compatible\end{tabular}}}} & \multicolumn{1}{c|}{\cellcolor[HTML]{EFEFEF}{\color[HTML]{333333} \textbf{\begin{tabular}[c]{@{}c@{}}Implementation\\   language\end{tabular}}}} & \multicolumn{1}{c|}{\cellcolor[HTML]{EFEFEF}{\color[HTML]{333333} \textbf{License}}} \\ \hline
QGroundControl & Free & Graphical & \begin{tabular}[c]{@{}l@{}}PX4 Pro, ArduPilot (APM)\\ or any vehicle that\\ communicates using\\ the MAVLink protocol.\end{tabular} & \begin{tabular}[c]{@{}l@{}}Windows/Mac/Linux/iOS\\ and Android devices\end{tabular} & Yes & C++ & \begin{tabular}[c]{@{}l@{}}Open Source\\ (GPLv3)\end{tabular} \\ \hline
Mission Planner & Free & Graphical & APM/PX4 & Windows/Mac OS (Using Mono) & Yes & C\# & \begin{tabular}[c]{@{}l@{}}Open source\\ (GPLv3)\end{tabular} \\ \hline
APM Planner  2.0 & Free & Graphical & \begin{tabular}[c]{@{}l@{}}MAVlink based\\ autopilots including\\ APM and PX4/Pixhawk\end{tabular} & Windows, Mac OS, and Linux & Yes & C++ & \begin{tabular}[c]{@{}l@{}}Open source\\ (GPLv3)\end{tabular} \\ \hline
MAVProxy & Free & \begin{tabular}[c]{@{}l@{}}Command line\\ and console\\ based interface\end{tabular} & \begin{tabular}[c]{@{}l@{}}Ardupilot MAVLink\\ compatible\end{tabular} & Linux & Yes & Python & \begin{tabular}[c]{@{}l@{}}Open source\\ (GPLv3)\end{tabular} \\ \hline
DroidPlanner & Free & Graphical & APM & Android Phones and Tablets & Yes & Java & \begin{tabular}[c]{@{}l@{}}Open source\\ (GPLv3)\end{tabular} \\ \hline
UGCS & \begin{tabular}[c]{@{}l@{}}Free version\\ with limited\\ capabilities\end{tabular} & Graphical & \begin{tabular}[c]{@{}l@{}}APM, Pixhawk, DJI,\\ Mikrokopter, YUNEEC,\\ Micropilot, Microunmanned systems,\\ Lokheed Martin, Parrot (Ar.unmanned system)\\ and other MAVLink compatible\\ multirotors, fixed wings\\ and VTOLs\end{tabular} & \begin{tabular}[c]{@{}l@{}}Windows, Mac OS,\\ Ubuntu, Android, iOS\end{tabular} & Yes & \begin{tabular}[c]{@{}l@{}}Human control\\ interface with C\#,\\ Universal control\\ server with JAVA,\\ Vehicle specific layer\\ with Java or C++\end{tabular} & \begin{tabular}[c]{@{}l@{}}Not open source with \\a free licence available\end{tabular} \\ \hline
\end{tabular}%
}
\label{tab:GCS}
\end{table*}

    \item \textbf{Mission Planner:} Is the second most popular ground stations for MAVLink-vehicles \cite{Mission}. It is created by Michael Oborne and runs on Windows platforms only. Similarly to  QGroundControl, Mission Planner also allows for planning an autonomous mission and making full control of the MAVLink vehicle. It has an additional feature of downloading and analyzing the log files of a mission. This means after completing any mission or operation of the unmanned systems, all internal parameters and state variables are stored in log files inside the autopilot, which can be download and analyzed by the Mission Planner. This helps you to understand how the autopilot behaves and provides a means to analyze any exotic behavior and analyze the performance of the autopilot. 
    
    \item \textbf{APM Planner 2.0:} Is also a ground station software that is very similar to mission planner but is also available for both MAC OS and Linux environments \cite{APM}. It is considered as the best ground station to use for MAC and Ubuntu operating systems. APM Planner 2.0 provides almost all the functionalities of Mission Planner including analyzing log files.

    \item \textbf{MAVProxy:} Is a Linux-based ground station, that is primarily a command line interface and console-based interface with some graphical modules for map visualization and mission editing \cite{MAVProxy}. MAVProxy is written in Python. It uses a set of simple command to interact with Ardupilot autopilot. The advantage of MAVProxy is that it is portable and lightweight as compared to other ground stations, and also quite easy to use. 

    \item \textbf{DroidPlanner:} For Android devices, DroidPlanner is also known as Tower software, is the best alternative for Android devices \cite{benemann2014droidplanner}. DroidPlanner relies on a Java ground station at a lower level that interfaces with the users through an Android GUI. It presents an excellent interface to interact with an autopilot through either serial telemetry interfaces and also network interfaces using both UDP and TCP like other ground stations. It also allows the user to configure the parameters of the autopilot and create missions on the fly. Nonetheless, it does not analyze log files like Mission Planner and APM Planner 2.
    
    \item \textbf{Universal Ground Control Software (UGCS):} It is a simple desktop software solution able to communicate with and control multiple unmanned systems simultaneously \cite{UGCS}. It also supports various autopilots from different manufacturers such as APM, Pixhawk, DJI, Mikrokopter, YUNEEC, Micropilot, Micro unmanned systems, Parrot (Ar.unmanned system ) and other MAVLink compatible. UGCS supports several map layers and map providers. It provides a much more robust interface with many features such as NFZs and immersive 3D simulation. It runs on different platforms, namely, Windows, Mac OS, Ubuntu, iOS and Android devices \cite{8450505}.

    
    
    
    
\end{itemize}

\begin{table*}[htbp]
\centering
\caption{Comparison between unmanned systems simulators}
\resizebox{\textwidth}{!}{%
\begin{tabular}{|c|l|l|l|l|l|l|l|l|l|l|}
\hline
\rowcolor[HTML]{EFEFEF} 
\textbf{Simulator} & \multicolumn{1}{c|}{\cellcolor[HTML]{EFEFEF}\textbf{Main domain}} & \multicolumn{1}{c|}{\cellcolor[HTML]{EFEFEF}\textbf{\begin{tabular}[c]{@{}c@{}}Commercial/\\ free\end{tabular}}} & \multicolumn{1}{c|}{\cellcolor[HTML]{EFEFEF}\textbf{\begin{tabular}[c]{@{}c@{}}Implementation\\  language\end{tabular}}} & \multicolumn{1}{c|}{\cellcolor[HTML]{EFEFEF}\textbf{Open source}} & \multicolumn{1}{c|}{\cellcolor[HTML]{EFEFEF}\textbf{\begin{tabular}[c]{@{}c@{}}Operating\\  systems\end{tabular}}} & \multicolumn{1}{c|}{\cellcolor[HTML]{EFEFEF}\textbf{License}} & \multicolumn{1}{c|}{\cellcolor[HTML]{EFEFEF}\textbf{\begin{tabular}[c]{@{}c@{}}Supported\\  Vehicles\end{tabular}}} & \multicolumn{1}{c|}{\cellcolor[HTML]{EFEFEF}\textbf{\begin{tabular}[c]{@{}c@{}}MAVLink\\  compatible\end{tabular}}} & \multicolumn{1}{c|}{\cellcolor[HTML]{EFEFEF}\textbf{\begin{tabular}[c]{@{}c@{}}ROS\\  interface\end{tabular}}} & \multicolumn{1}{c|}{\cellcolor[HTML]{EFEFEF}\textbf{SITL/HITL}} \\ \hline
\textbf{FlightGear} & unmanned systems & Free & C, C++ & Yes & \begin{tabular}[c]{@{}l@{}}Windows, Linux,\\ Mac OS-X, IRIX\\ FreeBSD, Solaris\end{tabular} & GNU/GPL & Aircraft, unmanned systems & Yes & No & Yes \\ \hline
\textbf{UE4Sim} & Vehicles & Free & Python, C++, & Yes &  &  & unmanned systems, cars & No & No & No \\ \hline
\textbf{X-Plane} & unmanned systems & Commercial & C++ & No & \begin{tabular}[c]{@{}l@{}}Android, iOS, Linux,\\ MacOS, WebOS,\\ Windows\end{tabular} & \begin{tabular}[c]{@{}l@{}}Proprietary with\\ Free Trial\end{tabular} & Plane & Yes & No & HITL \\ \hline
\textbf{AirSim} & unmanned systems, cars & Free & \begin{tabular}[c]{@{}l@{}}C++, Python,\\ C\#, Java\end{tabular} & Yes & Windows, Linux & MIT & \begin{tabular}[c]{@{}l@{}}Iris (MultiRotor model\\ and a configuration\\ for PX4 QuadRotor in\\ the X configuration)\end{tabular} & Yes & No & Yes \\ \hline
\textbf{Gazebo} & Robots & Free & \begin{tabular}[c]{@{}l@{}}C++,\\ JavaScript\end{tabular} & Yes & \begin{tabular}[c]{@{}l@{}}Linux, Mac\\ Windows\end{tabular} & Apache V2.0 & \begin{tabular}[c]{@{}l@{}}Quad ( Iris and Solo),\\ Hex (Typhoon H480), \\ Generic quad delta VTOL,\\ Tailsitter, Plane, Rover,\end{tabular} & Yes & Yes & Yes \\ \hline
\textbf{jMAVSim} & unmanned systems & Free & JAVA & Yes & \begin{tabular}[c]{@{}l@{}}Linux, MacOs,\\  Windows\end{tabular} & BSD 3 & Multirotor/Quad & Yes & Yes & Yes \\ \hline
\end{tabular}%
}
\label{tab:unmanned system}
\end{table*}

\subsection{Unmanned Systems Simulators}
Unmanned systems simulator help simulating any environment and any unmanned systems activities in a digital environment to make easier the test and the validation of algorithms and protocols developed for the UAVs. The choice of the appropriate simulator depends on the objectives, the areas of application and the functionalities given by the simulator. Table \ref{tab:unmanned system} provides a comparison between unmanned systems simulators discussed in this section.

\begin{itemize}

    \item \textbf{FlightGear:} Is a free, open-source flight simulator framework used for research and academic environments \cite{perry2004flightgear}. It works on different environment such as Windows, Mac, and Linux operating systems platforms \cite{8282877}. The entire source code is available for modification and published under the General Public License (GPL). Aircraft models must be created by an external 3D modeling application. Typically, the UAV structure and features are described by an XML file. FlightGear can run Software-In-The-Loop (SITL) and Hardware-In-The-Loop (HITL).
    \item \textbf{UE4Sim:} In 2017, the Unreal Engine (UE4) is developed at King Abdullah University of Science and Technology, based on the open-source computer game engine Unreal Engine (UE4) \cite{mueller2017ue4sim}. The simulator has been designed to facilitate the integration of computer vision and machine learning techniques into a realistic looking 3D environment. UE4Sim gives accurate unmanned system physics, an evaluation tool based on the latest advanced tracking algorithms, and a deep learning interface based on TensorFlow for autonomous driving without requiring manually collected training data.

    \item \textbf{X-Plane:} Is a commercial flight simulator produced by Laminar Research \cite{xplane}. X-Plane simulator works on different environment, namely, Windows, Linux, and Android. It is certified by the FAA (Federal Aviation Administration) as a training simulator because it is more flexible and offers high fidelity simulation than the flight model when it is used with specific hardware configurations \cite{Garcia2010}.
    The flight model was created using the Plane-Maker, an application provided with X-Plane. This tool allows users to design any aircraft based on the vehicle's physical specifications.
    X-Plane uses UDP or TCP-based protocols to connect different instances through a network. X-Plane can exchange information through the UDP communication protocol, which guarantees high-speed data traffic.

    \item \textbf{Aerial Informatics and Robotics Platform (AirSim):} This simulator was produced in 2017 by Microsoft to develop and test deep learning, computer vision, and reinforcement learning algorithms for autonomous vehicle
applications \cite{shah2018airsim}. It is open-source, cross-platform, built on Unreal Engine 4 (UE4), and supports SITL and HITL with popular flight controllers such as Ardupilot and PX4 with the possibility of interfacing with MAVLink protocol to render the simulation more realistic \cite{air-sim}.

AirSim can retrieve data, images, control and interact with the vehicle based on APIs, via C++, Python, C\# and Java languages. However, AirSim simulation is limited to quadunmanned systems. AirSim does not support ROS and cloud connectivity \cite{Johansen2017}. This simulator is also computation-extensive and needs advanced computing requirements as compared to other simulators. 
 \item \textbf{Gazebo:} Is an open source simulation tool for robots and vehicles used for several applications \cite{Koenig04}. This simulator was developed at the University of Southern California and currently managed by the Open Source Robotics Foundation (OSRF). 
It supports different robots and can simulate complex 3D virtual worlds with supporting various physical simulation engines and different sensors, to test robot designs and AI algorithms using real scenarios.
Gazebo supports Ardupilot and PX4 with the ability to run Software In The Loop and Hardware-In-Tl-Loop. An API is provided allowing the creation of new sensors for Gazebo.
 Moreover, Gazebo is one of the most popular simulators since it enables multi-robot simulation,  supports ROS and enables cloud connectivity \cite{Gazebo}. However, it is too computationally demanding to simulate multi-vehicle operations in real-time \cite{read2013profiling}.
    \item \textbf{Java Micro Air Vehicle Simulator (jMAVSim):} Is a Java UAV simulator developed by the PIXHAWK engineering team \cite{jmavsim}. The main advantage of jMAVSim is that it is simple to use and lighweight.     jMAVSim allows flying unmanned system type vehicles running PX4 around a simulated world. It supports the MAVLink protocol with the possibility to run SITL via UDP and HITL via a serial connection. It supports ROS and uses a Java3D library for graphical visualization. There is no possibility to integrate other sensors in the simulation.

\end{itemize}


\section{CONCLUSIONS} \label{sec:con}
In this paper, we presented a comprehensive survey on the MAVLink protocol, which is a lightweight protocol for communication with unmanned systems. This survey addresses the need for having a technical reference for MAVLink-based systems developers. We thorough presented the characteristics of the MAVLink protocol version 1 and version 2 and their messages formats. Furthermore, we discussed the MAVLink security requirements, threats, possible solutions, and we presented the recent research works that dealt with security aspects of MAVLink. We also presented a comprehensive literature review of related research works about MAVLink. 

We believe that this survey provides a handy reference for the large community of practitioners and developers to learn about the MAVLink protocol, in particular with the absence of any technical coverage of MAVLink except some online documentation resources.

\section*{Acknowledgments}
This work is supported by the Robotics and Internet of Things Lab of Prince Sultan University, and Jinan University. 

\bibliographystyle{IEEEtran}
\bibliography{IEEEabrv,biblio}

\begin{IEEEbiographynophoto}{\textbf{ANIS KOUB\^AA}} received the M.Sc. degree from University Henri Poincar\'e, France, in 2001, and the Ph.D. degree from INPL, France, in 2004. He is currently a Professor of computer science, Aide to Rector of Research Governance, and the Director of the Robotics and Internet of Things Research Lab, Prince Sultan University. He is also a Senior Researcher with CISTER/INESC and ISEP-IPP, Porto, Portugal, and a Research and Development Consultant with Gaitech Robotics, China. His current research interests include providing solutions towards the integration of robots and drones into the Internet of Things (IoT) and clouds, in the context of cloud robotics, robot operating systems, robotic software engineering, wireless communication for the IoT, real-time communication, safety and security for cloud robotics, intelligent algorithms design for mobile robots, and multi-robot task allocation. He is also a Senior Fellow of the Higher Education Academy, U.K. He has been the Chair of the ACM Chapter, Saudi Arabia, since 2014.

\end{IEEEbiographynophoto}

\begin{IEEEbiographynophoto}{\textbf{AZZA ALLOUCH}} was born in Tunisia, in 1991. She received the master’s degree from the National School of Electronics and Telecommunication, Sfax, in 2016. She is currently pursuing the Ph.D. degree with the School of Intelligent Systems Science and Engineering, Jinan University, China, in cooperation with the Faculty of Mathematical, Physical and Natural Sciences of Tunis (FST), University of El Manar, Tunisia. She is currently an Associate Member of the LISI Laboratory, National Institute of Applied Sciences and Technology, University of Carthage, Tunisia. She is also with the Robotics and Internet-of-Things Laboratory, Prince Sultan University, Saudi Arabia, and Gaitech Robotics, China. Her research interests include machine learning, Blockchain, and UAV.

\end{IEEEbiographynophoto}

\begin{IEEEbiographynophoto}{\textbf{MARAM ALAJLAN}} is currently a Ph.D. student in Computer Science at King Saud University, Saudi Arabia. She received her M.Sc. and B.Sc. degrees in Computer Science from Al-Imam Muhammad Ibn Saud Islamic University, Saudi Arabia, in 2014 and 2009, respectively. Her research interests are mobile robots and cloud computing.

\end{IEEEbiographynophoto}

\begin{IEEEbiographynophoto}{\textbf{YASIR JAVED }} is a skilled senior programmer / developer with more than fourteen years’ of experience in programming, software development, project management and analytics. His research interest includes Programming, Robotics, Drones, Vehicular Platoons, Secure Software development, Mobile Apps Security, Signal processing, IoT Analytics, Intelligent Applications, Statistics, data analytics, Forensics Analysis, big data and Predictive computing. He is also working as Research Engineer at COINS research group.  He has successfully completed various International and National Research funding projects and has served as Analyst programmer at Prince Megren Data Center, Center of Excellence and Research and initiative center at Prince Sultan University. He is currently an active member of RIOTU group at Prince Sultan University.

\end{IEEEbiographynophoto}

\begin{IEEEbiographynophoto}{\textbf{ABDELFETTAH BELGHITH }} received his Master of Science and his PhD degrees in computer science from the University of California at Los Angeles (UCLA) respectively in 1982 and 1987. He is since 1992 a full Professor at the National School of Computer Sciences (ENSI), University of Manouba, Tunisia. He is currently on a sabbatical leave at King Saud University, Saudi Arabia. His research interests include computer networks, wireless networks, multimedia Internet, mobile computing, distributed algorithms, systems and information security, simulation and performance evaluation. He runs several research projects in cooperation with other universities, research laboratories and research institutions. He is the Past chair of the IEEE Tunisia section, the chair of the IEEE ComSoc and VTS Tunisia Chapters, and the Director of the HANA Research Laboratory (www.hanalab.org) at the National School of Computer Sciences. He published more than 350 research papers in international journals and conference proceedings.

\end{IEEEbiographynophoto}

\begin{IEEEbiographynophoto}{\textbf{MOHAMED KHALGUI}} received the B.S. degree in computer science from Tunis El Manar University, Tunis, Tunisia, in 2001, the M.S. degree in telecommunication and services from Henri Poincar\'e University, Nancy, France, in 2003, the Ph.D. degree in computer science from the National Polytechnic Institute of Lorraine, Nancy, France, in 2007, and the Habilitation Diploma in information technology (computer science) from the Martin Luther University of Halle-Wittenberg, Halle, Germany, in 2012. He was a Researcher in computer science with the Institut National de Recherche en Informatique et Automatique (INRIA), Rocquencourt, France, the ITIA-CNR Institute, Vigevano, Italy, and Systems Control Laboratory, Xidian University, Xi'an, China. He was a Collaborator with the KACST Institute, Riyadh, Saudi Arabia, the SEG Research Group, Patras University, Patras, Greece, and the Director of the RECS Project of O3NEIDA, Ottawa, ON, Canada. He was the Manager of the RES Project of the Synesis Consortium, Lomazzo, Italy, the Cyna-RCS Project of the Cynapsys Consortium, Paris, France, and the BROS and RWiN Projects of the ARDIA Corporation, Tunis, Tunisia. He was a Professor with the National Institute of Applied Sciences and Technology, University of Carthage, Tunisia. He is currently a Professor with Jinan University, Zhuhai, China. He has been involved in various international projects and collaborations. Dr. Khalgui is a member of different conferences and boards of journals. He received the Humboldt Grant in Germany for preparing the Habilitation Diploma at the Martin Luther University of Halle-Wittenberg. He is nominated Talented Full Professor by Chinese government in 2019.    
\end{IEEEbiographynophoto}


\end{document}